\documentclass{article}

\usepackage[preprint]{neurips_2025}

\usepackage[utf8]{inputenc} 
\usepackage[T1]{fontenc}    
\usepackage{hyperref}       
\usepackage{url}            
\usepackage{booktabs}       
\usepackage{amsfonts}       
\usepackage{nicefrac}       
\usepackage{microtype}      
\usepackage{xcolor}         
\usepackage{graphicx}
\usepackage{amsmath}
\usepackage{amsfonts}
\usepackage{multirow}
\usepackage{subfigure}
\usepackage{pgfplots}
\usepackage{xspace}
\usepackage{enumitem}
\usepackage{wrapfig}
\usepackage{bbding}
\usepackage{colortbl}
\usepackage{adjustbox}
\usepackage{caption}

\title{Exploring and Exploiting the Inherent Efficiency within Large Reasoning Models for Self-Guided Efficiency Enhancement}

\author{%
  Weixiang Zhao\textsuperscript{1}, Jiahe Guo\textsuperscript{1}, Yang Deng\textsuperscript{2}, \textbf{Xingyu Sui\textsuperscript{1}}, \textbf{Yulin Hu}\textsuperscript{1} \\  \textbf{Yanyan Zhao\textsuperscript{1}}, \textbf{Wanxiang Che\textsuperscript{1}}, \textbf{Bing Qin\textsuperscript{1}}, \textbf{Tat-Seng Chua\textsuperscript{3}}, \textbf{Ting Liu\textsuperscript{1}}\\
  \textsuperscript{1}Harbin Institute of Technology, \textsuperscript{2}Singapore Management University \\
  \textsuperscript{3}National University of Singapore \\
  \texttt{\{wxzhao,jhguo,yyzhao\}@ir.hit.edu.cn} \\
}

\begin{document}

\maketitle

\begin{abstract}
  Recent advancements in large reasoning models (LRMs) have significantly enhanced language models’ capabilities in complex problem-solving by emulating human-like deliberative thinking. However, these models often exhibit overthinking (\textit{i.e.}, the generation of unnecessarily verbose and redundant content), which hinders efficiency and inflates inference cost. In this work, we explore the representational and behavioral origins of this inefficiency, revealing that LRMs inherently possess the capacity for more concise reasoning. Empirical analyses show that correct reasoning paths vary significantly in length, and the shortest correct responses often suffice, indicating untapped efficiency potential.
Exploiting these findings, we propose two lightweight methods to enhance LRM efficiency. First, we introduce Efficiency Steering, a training-free activation steering technique that modulates reasoning behavior via a single direction in the model’s representation space. Second, we develop Self-Rewarded Efficiency RL, a reinforcement learning framework that dynamically balances task accuracy and brevity by rewarding concise correct solutions. Extensive experiments on seven LRM backbones across multiple mathematical reasoning benchmarks demonstrate that our methods significantly reduce reasoning length while preserving or improving task performance. Our results highlight that reasoning efficiency can be improved by leveraging and guiding the intrinsic capabilities of existing models in a self-guided manner. 
\end{abstract}

\section{Introduction}

Recent progress in large language models (LLMs), notably OpenAI's o1/o3/o4 models \citep{jaech2024openai,openai2025o3blog} and the DeepSeek-R1 series \citep{guo2025deepseek}, has marked a clear shift toward the development of large reasoning models (LRMs). Compared to conventional LLMs \citep{brown2020language,dubey2024llama,team2024gemma,yang2024qwen2}, LRMs excel at solving complex reasoning problems by emulating human-like deliberative thinking. A defining characteristic of LRMs is their capacity for extensive chain-of-thought reasoning, whereby they generate structured reasoning traces before arriving at a final answer \citep{li2025system,xu2025towards,chen2025towards}.
Although LRMs have achieved remarkable advances in reasoning abilities, a critical issue of overthinking has become increasingly apparent, leading to substantial inefficiencies \citep{chen2024not,ballon2025relationship}. These inefficiencies are primarily reflected in the generation of redundant or unnecessary content, such as repeated rephrasings, verbose justifications, or excessive analysis of otherwise simple problems. This not only inflates the response length but also hinders the overall efficiency and user experience of LRM-based systems.

However, mitigating overthinking is particularly challenging, as it is generally impossible to predict in advance how efficiently and effectively a given problem can or should be solved \citep{team2025kimi,arora2025training,aggarwal2025l1,qu2025optimizing,hou2025thinkprune}. Reasoning complexity varies widely across inputs, and rigid constraints on response length risk suppressing legitimate, necessary reasoning for harder cases \citep{ma2025cot,munkhbat2025self,kang2025c3ot,liu2024can}. Furthermore, the underlying mechanisms behind LRMs' inefficiency remain largely unexplored, and the community currently lacks a clear understanding of why overthinking emerges during generation. Therefore, any solution must carefully balance brevity and adequacy, ideally in a way that adapts dynamically to the problem at hand, without relying on external supervision or handcrafted heuristics.

Motivated by this issue, we begin by examining whether such inefficiency is an inherent and avoidable property within LRMs themselves (\S\ref{sec:lrm_efficiency}). Specifically, when sampling multiple responses to the same problem, we consistently observe that the shortest correct solution is often less than half the length of the longest correct one. This pattern holds across LRMs with diverse training paradigms, including RL-optimized (e.g., QwQ \citep{qwen2025qwqblog}, GLM-Z1 \citep{glm2024chatglm}) and distilled variants (e.g., DeepSeek-Qwen-Distill \citep{guo2025deepseek}). These findings reveal a surprising degree of potential efficiency embedded within LRMs.
To gain deeper insight into the origins of this intrinsic efficiency, we explore its underlying mechanisms from a representational standpoint (\S\ref{sec:deep_analysis}). Within the representation space, efficient reasoning paths consistently exhibit distinct positional shifts from their more verbose counterparts across all layers of the LRM, with such deviations clearly observable across problems of varying difficulty levels. This subtle divergence effectively minimizes unnecessary self-reflective behavior, as reflected by a reduced occurrence of indicative expressions such as \textit{“Wait”} and \textit{“Alternatively”}, along with a lower prevalence of \textit{Reflection} and \textit{Transition} phases in the model’s reasoning process. These findings reveal clear representational, lexical and behavioral distinctions between efficient and inefficient reasoning trajectories.

To this end, the representational and behavioral insights discussed above motivate two methods for enhancing reasoning efficiency in LRMs through self-guided mechanisms: one based on direct training-free intervention in the model's internal representations, and the other grounded in reinforcement learning with efficiency-aware rewards.

On one hand, leveraging representational findings, we introduce a lightweight, training-free method called \emph{Efficiency Steering}, which activates the model’s intrinsic capacity for efficient reasoning. We show that a single vector direction in the representational space, which is derived from the positional shifts between efficient and verbose reasoning trajectories across layers, can effectively modulate reasoning behavior. This vector is computed using a small set of contrastive reasoning pairs and captures the key directional difference in their internal dynamics. By applying this direction during inference, we can steer the model toward either more concise or more elaborate reasoning paths.

On the other hand, the lexical patterns observed in efficient reasoning motivate a reinforcement learning strategy, named \emph{Self-Rewarded Efficiency RL}, which explicitly incentivizes concise reasoning. This approach incorporates a reward function that balances task accuracy with a dynamic length penalty, calibrated against the shortest successful reasoning trace encountered in each rollout. As a result, the model is guided to produce solutions that are not only correct but also succinct.

Extensive experiments conducted across 7 different LRMs, with maximum parameter scales up to 32B parameters, validate 
the effectiveness of the proposed Efficiency Steering and Self-Rewarded Efficiency RL for efficiency enhancement. The experimental results clearly demonstrate that our two methods succeed to enhance reasoning efficiency on four mathematical reasoning benchmarks without compromising overall performance \citep{lightman2023let,maa_amc}.

Overall, our study reveals that LRMs already possess a substantial degree of untapped reasoning efficiency, which can be surfaced and guided through simple representational interventions and self-rewarded regulation. This suggests that improving model efficiency does not always require external supervisions or complex optimization pipelines. Instead, a better understanding of the model's internal behaviors may open up lightweight and practical alternatives. We hope this work encourages future research to explore and refine the capabilities that already exist within LRMs for building more efficient and accessible reasoning systems.

\begin{figure*}
  \centering
  \begin{minipage}[t]{0.48\textwidth}
    \centering
    \includegraphics[width=\linewidth]{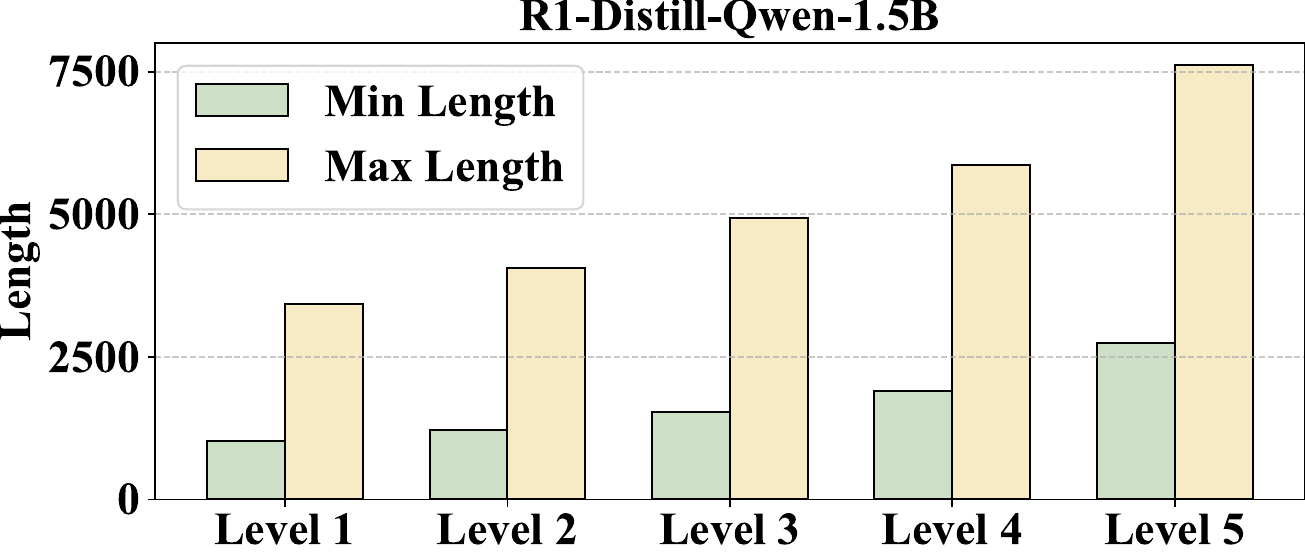}
    \caption*{(a) Comparison of the shortest and longest reasoning path lengths on the R1-Distill-Qwen-1.5B.}
  \end{minipage}
  \hfill
  \begin{minipage}[t]{0.48\textwidth}
    \centering
    \includegraphics[width=\linewidth]{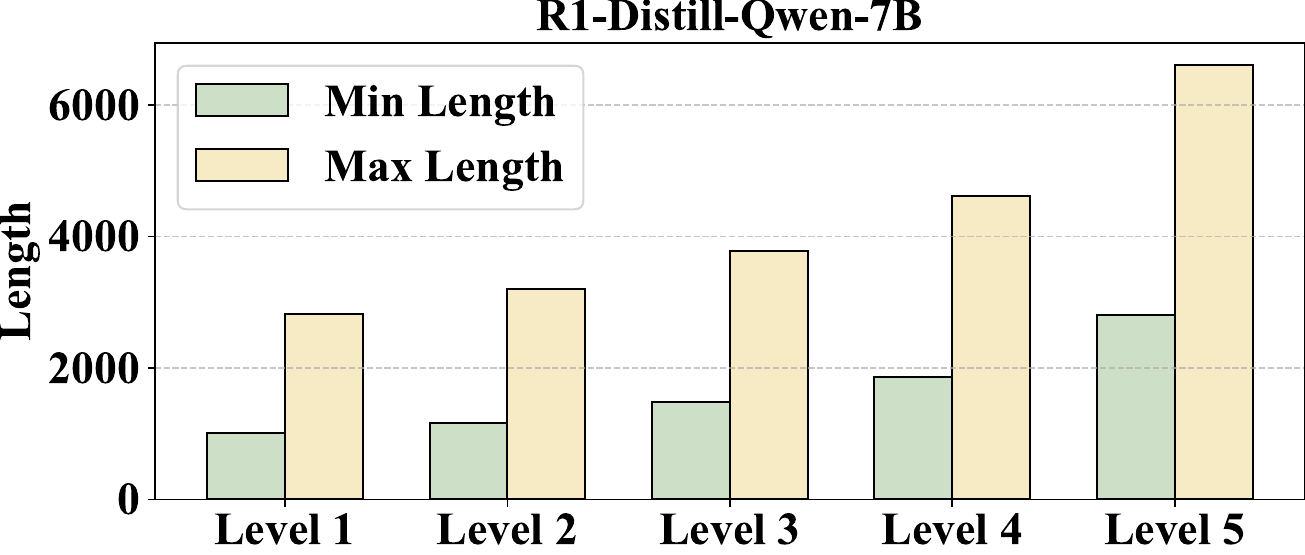}
    \caption*{(b) Comparison of the shortest and longest reasoning path lengths on the R1-Distill-Qwen-7B.}
  \end{minipage}
  \hfill \\
  \begin{minipage}[t]{0.48\textwidth}
    \centering
    \includegraphics[width=\linewidth]{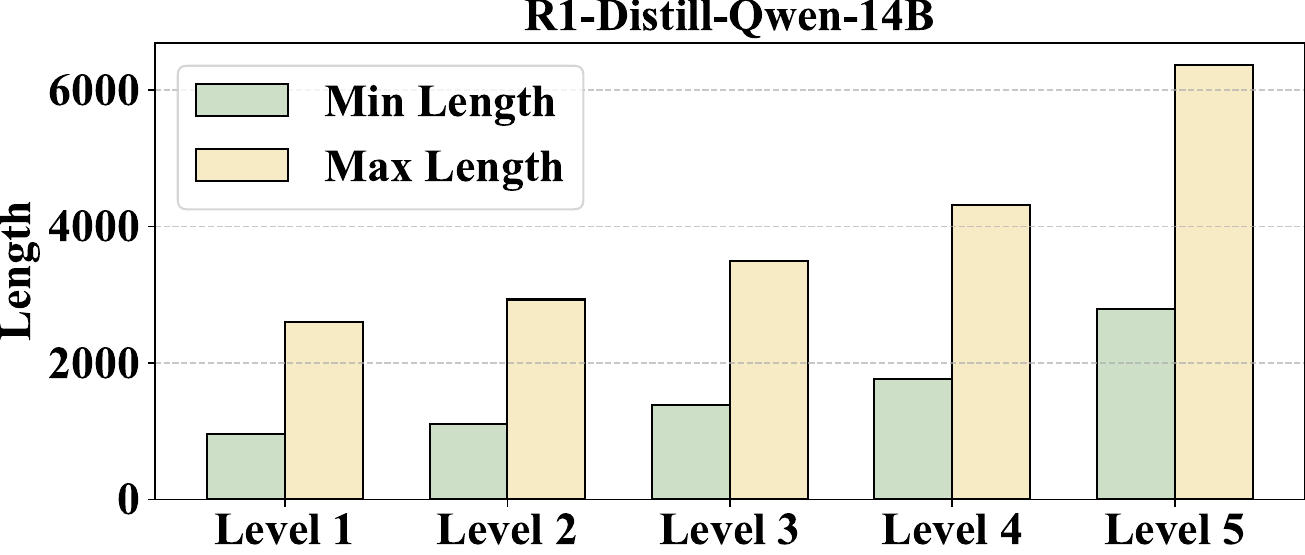}
    \caption*{(c) Comparison of the shortest and longest reasoning path lengths on the R1-Distill-Qwen-14B.}
  \end{minipage}
  \hfill
  \begin{minipage}[t]{0.48\textwidth}
    \centering
    \includegraphics[width=\linewidth]{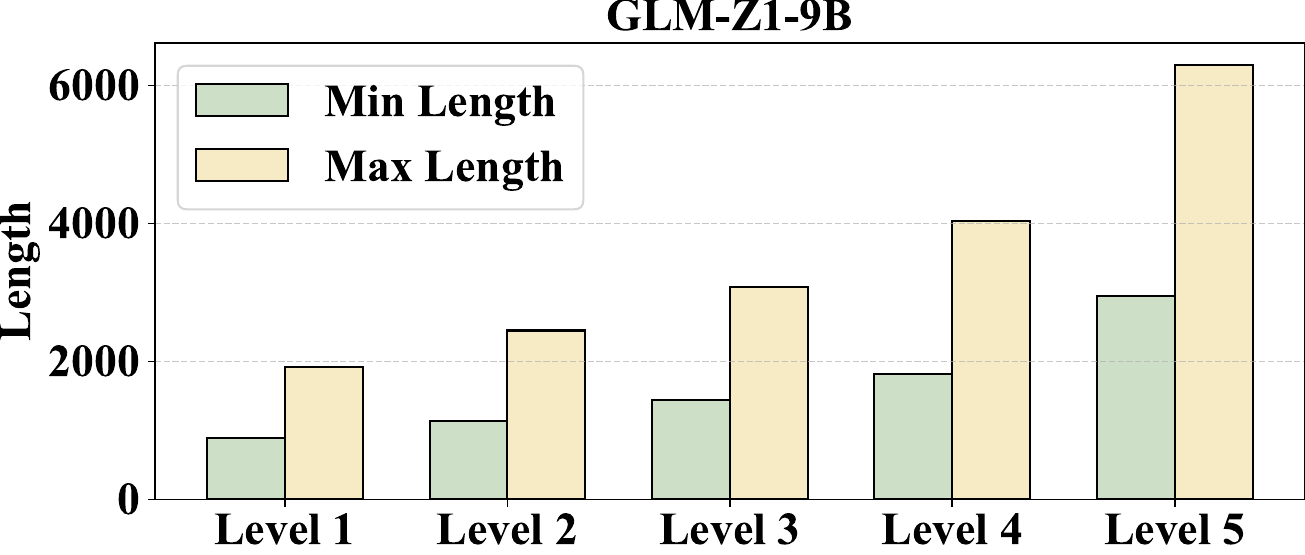}
    \caption*{(d) Comparison of the shortest and longest reasoning path lengths on the GLM-Z1-9B.}
  \end{minipage}
  \hfill
  \begin{minipage}[t]{0.48\textwidth}
    \centering
    \includegraphics[width=\linewidth]{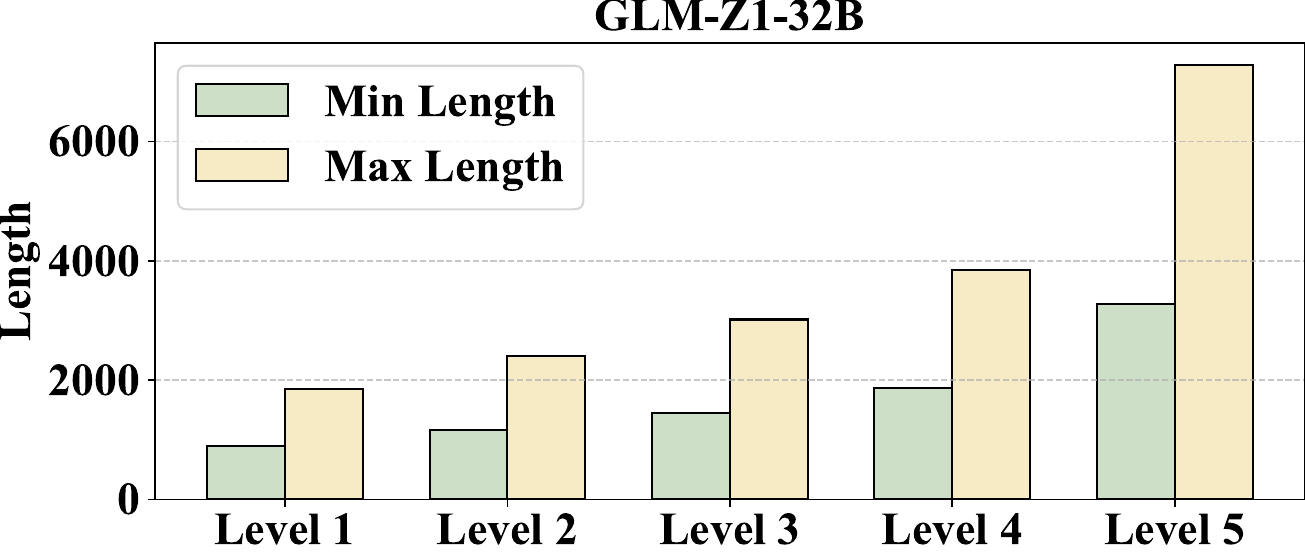}
    \caption*{(e) Comparison of the shortest and longest reasoning path lengths on the GLM-Z1-32B.}
  \end{minipage}
  \begin{minipage}[t]{0.48\textwidth}
    \centering
    \includegraphics[width=\linewidth]{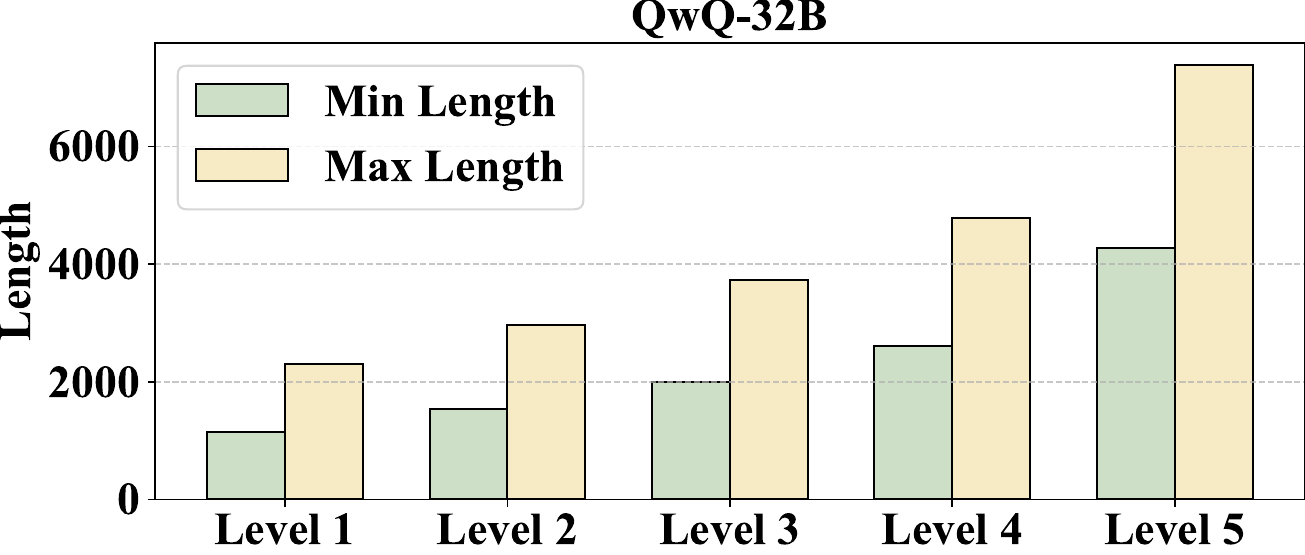}
    \caption*{(f) Comparison of the shortest and longest reasoning path lengths on the QwQ-32B.}
  \end{minipage}
  \hfill \\
  \caption{Comparison of the shortest and longest reasoning path lengths on different LRMs, including (a) R1-Distill-Qwen-1.5B, (b) R1-Distill-Qwen-7B, (c) R1-Distill-Qwen-14B, (d) GLM-Z1-9B, (e) GLM-Z1-32B and (f) QwQ-32B.}
  \label{fig:min_max_length_main}
\end{figure*}

\section{LRMs Naturally Exhibit Potential for Greater Efficiency}
\label{sec:lrm_efficiency}

In this section, we explore and demonstrate
the inherent capabilities of large reasoning models (LRMs) to achieve higher efficiency in mathematical reasoning tasks.

\begin{table}
    \centering
    \small
    \caption{This table summarizes the LRMs of different model families and scales evaluated for foundational capabilities and the fine-tuned source model.}
    \label{tab:model-overview}
    \resizebox{\linewidth}{!}{
    \begin{tabular}{l l  l  l  l}
    \toprule
        \bf Method & \bf LRM & \textbf{Fine-tuned Model} & \bf Ref. \\
        \midrule
        \multirow{4}{*}{Distillation} & R1-Distill-Qwen-1.5B & Qwen2.5-Math-1.5B & \citep{guo2025deepseek} \\ 
        ~ & R1-Distill-Qwen-7B & Qwen2.5-Math-7B & \citep{guo2025deepseek} \\ 
        ~ & R1-Distill-Qwen-14B & Qwen2.5-14B & \citep{guo2025deepseek} \\ 
        ~ & R1-Distill-Llama-8B & Llama-3.1-8B & \citep{guo2025deepseek} \\
        \midrule
        \multirow{3}{*}{Large-Scale RL} & GLM-Z1-9B & - & \citep{glm2024chatglm} \\
        ~ &GLM-Z1-32B &- & \citep{glm2024chatglm} \\
        ~ &QwQ-32B &- & \citep{qwen2025qwqblog} \\
        \bottomrule
    \end{tabular}}
\end{table}

\paragraph{Models} We conduct a comprehensive analysis of 7 LRMs from diverse model families and across a range of scales, obtained either through distillation or large-scale reinforcement learning. Our goal is to systematically explore the potential efficiency inherent in these models. Specifically, we examine models ranging from 1.5B to 32B parameters, covering the DeepSeek \citep{guo2025deepseek}, Qwen \citep{yang2024qwen2}, LLaMA \citep{dubey2024llama}, and GLM \citep{glm2024chatglm} families. All models included in the study are specified in Table \ref{tab:model-overview}.

\paragraph{Datasets} We use the 7,500 training sample prompt set of MATH \citep{hendrycks2021measuring}, which provides verifiable ground truth answers. This dataset comprises mathematics problems categorized into 5 difficulty levels. For each prompt, we sample a fixed number 8 of candidate responses and subsequently filter to retain only those whose final answers match the provided ground truth answers. From these retained correct reasoning paths, we compute the lengths of the shortest and longest reasoning paths. By default, model outputs are generated using a temperature setting of $t = 0.6$, a top-$p$ value of 0.95, and a maximum token output limit of 32,768.

\paragraph{Results and Analysis} Figure \ref{fig:min_max_length_main} presents the shortest and longest lengths of \emph{correct} reasoning paths generated by the above representative LRMs, across the five difficulty levels of the MATH dataset. We draw the key insight:

\textbf{LRMs inherently demonstrate strong potential for efficient reasoning.} As the problem difficulty increases, all models naturally generate longer reasoning traces, reflecting the increasing complexity of the tasks. However, across all difficulty levels, we observe a pronounced gap between the shortest and longest successful reasoning paths---often exceeding a 2× difference in token usage. For instance, in Level 5, R1-Distill-Qwen-7B shows a minimum path length of approximately 3,000 tokens, compared to over 6,000 tokens for the maximum path; similarly, GLM-Z1-9B reaches nearly 6,500 tokens at its longest, while maintaining a minimum under 3,500. This pattern holds consistently across QwQ-32B.

These findings indicate that even in the absence of any explicit efficiency supervision, LRMs are capable of generating significantly more concise reasoning traces—often reducing token usage by more than half. This latent ability suggests that LRMs already possess the internal capacity for efficient problem solving, and that surfacing this potential through targeted guidance (e.g., representation steering or efficiency-aware reward) is both feasible and impactful.

\section{Deeper Analysis on the Inherent Efficiency within LRMs}
\label{sec:deep_analysis}

In this section, we investigate the origins of the inherent efficiency observed in LRMs by conducting a twofold analysis. First, we examine representation-level dynamics (\S\ref{subsec:repre_analysis}), identifying consistent positional shifts in hidden states between efficient and verbose reasoning paths across layers. Second, we explore behavioral-level correlates (\S\ref{subsec:lex_analysis}), revealing how these representational differences manifest in reasoning style. Together, these analyses offer a comprehensive view of how inherent efficiency is internally encoded and behaviorally expressed within LRMs.

\begin{figure*}
  \centering
  \begin{minipage}[t]{0.32\textwidth}
    \centering
    \includegraphics[width=\linewidth]{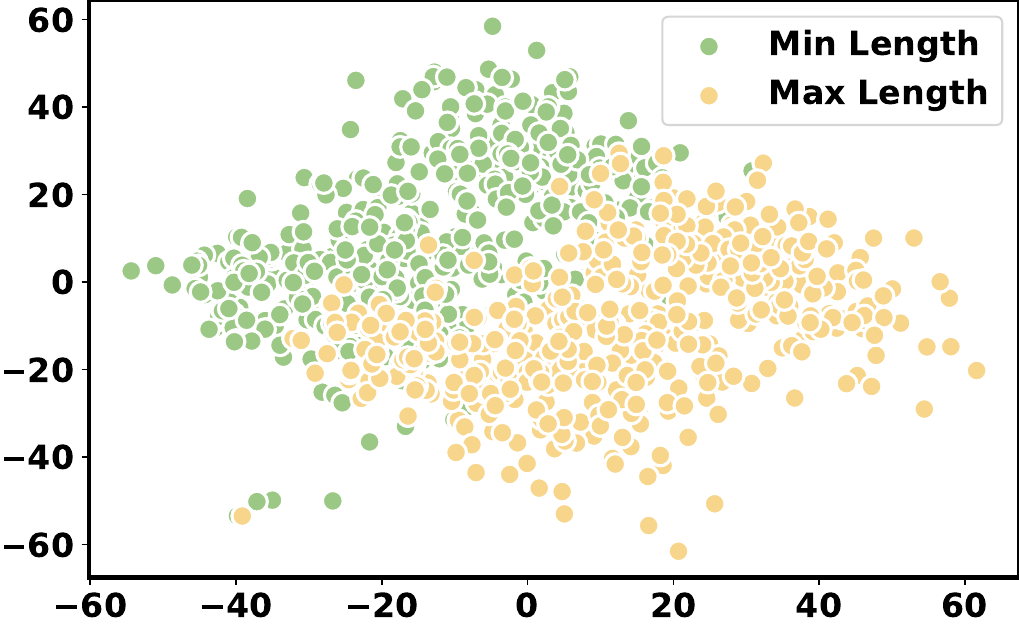}
    \caption*{(a) Layer 25 on MATH Level 1.}
  \end{minipage}
  \begin{minipage}[t]{0.32\textwidth}
    \centering
    \includegraphics[width=\linewidth]{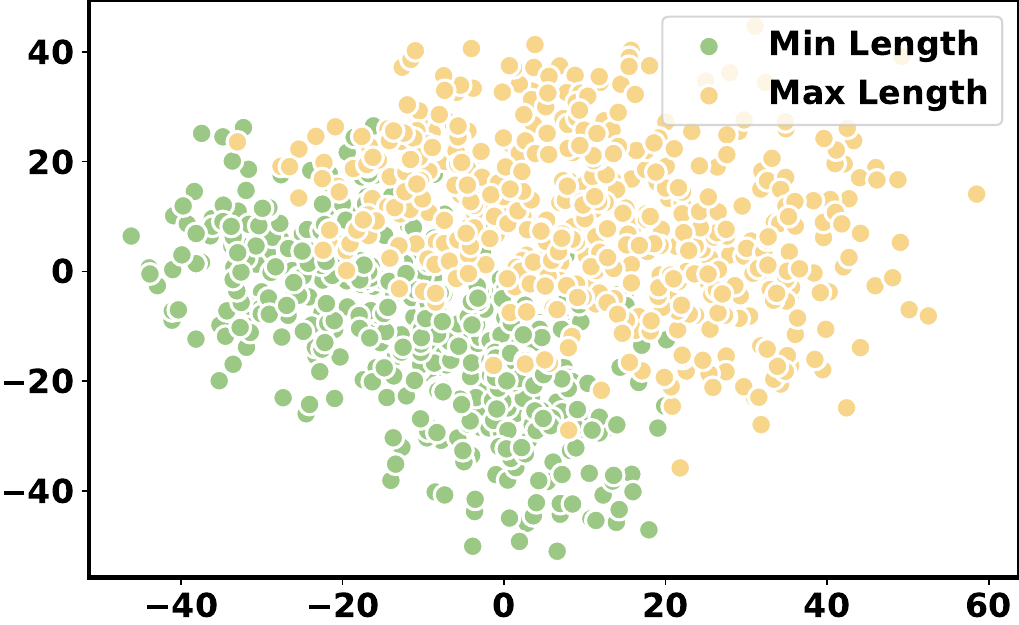}
    \caption*{(b) Layer 25 MATH Level 3.}
  \end{minipage}
  \begin{minipage}[t]{0.32\textwidth}
    \centering
    \includegraphics[width=\linewidth]{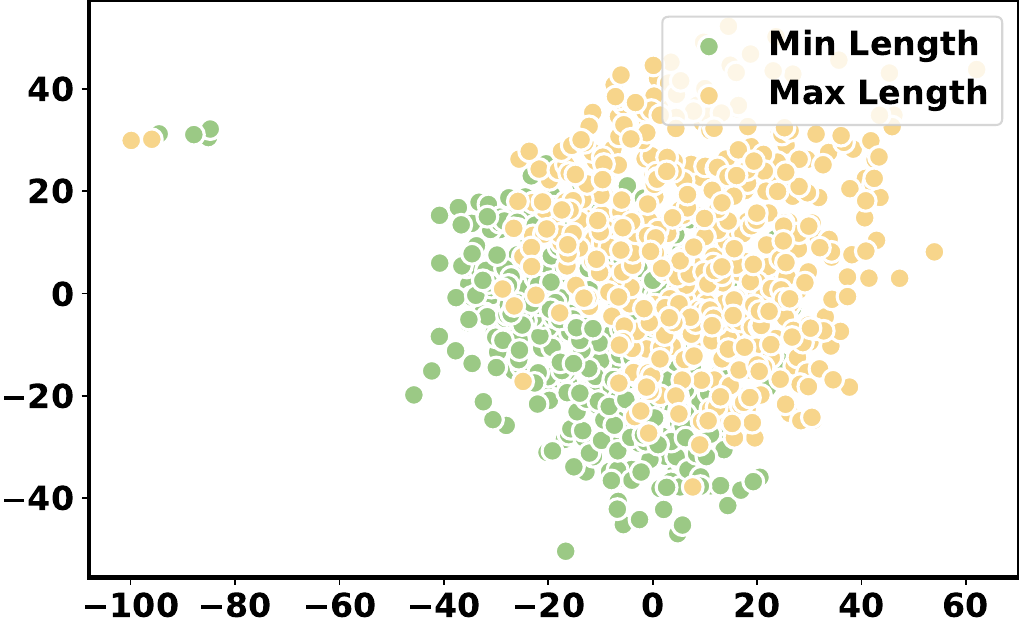}
    \caption*{(c) Layer 25 on MATH Level 5.}
  \end{minipage} \\
  \begin{minipage}[t]{0.32\textwidth}
    \centering
    \includegraphics[width=\linewidth]{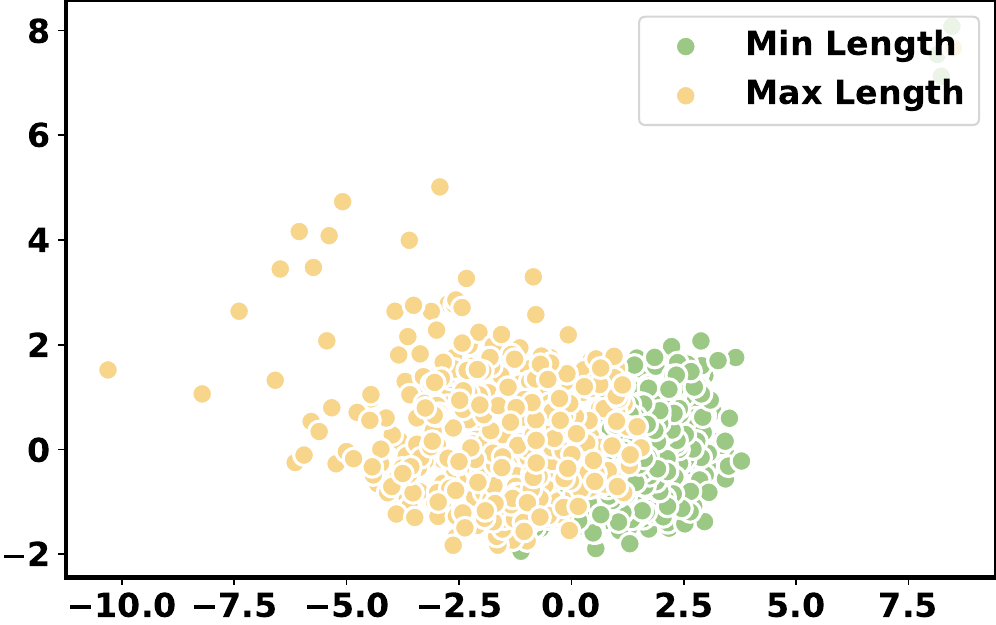}
    \caption*{(d) Layer 5 on MATH Level 3.}
  \end{minipage}
  \begin{minipage}[t]{0.32\textwidth}
    \centering
    \includegraphics[width=\linewidth]{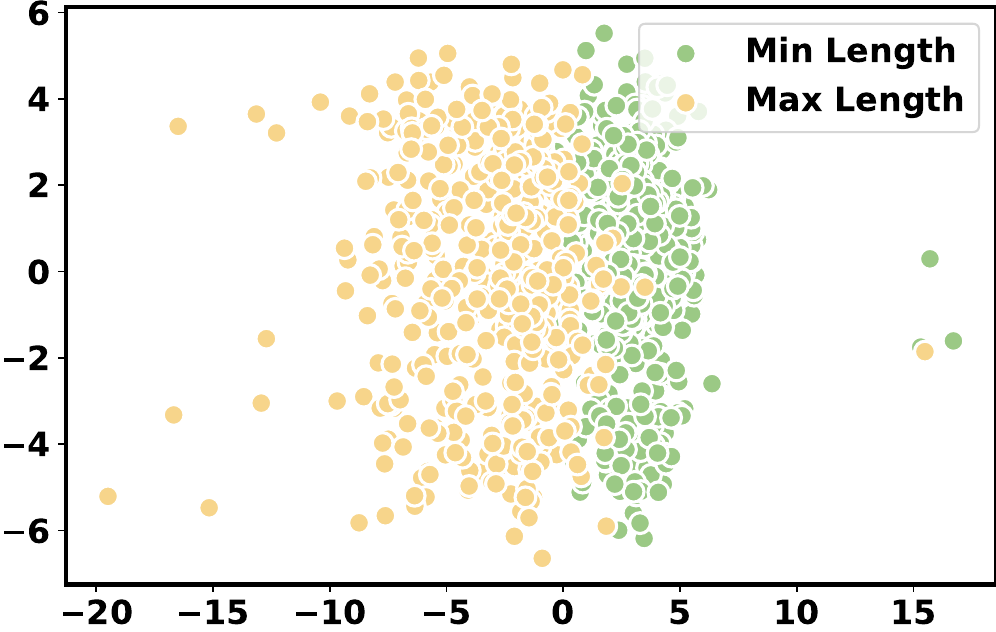}
    \caption*{(e) Layer 10 on MATH Level 3.}
  \end{minipage}
  \begin{minipage}[t]{0.32\textwidth}
    \centering
    \includegraphics[width=\linewidth]{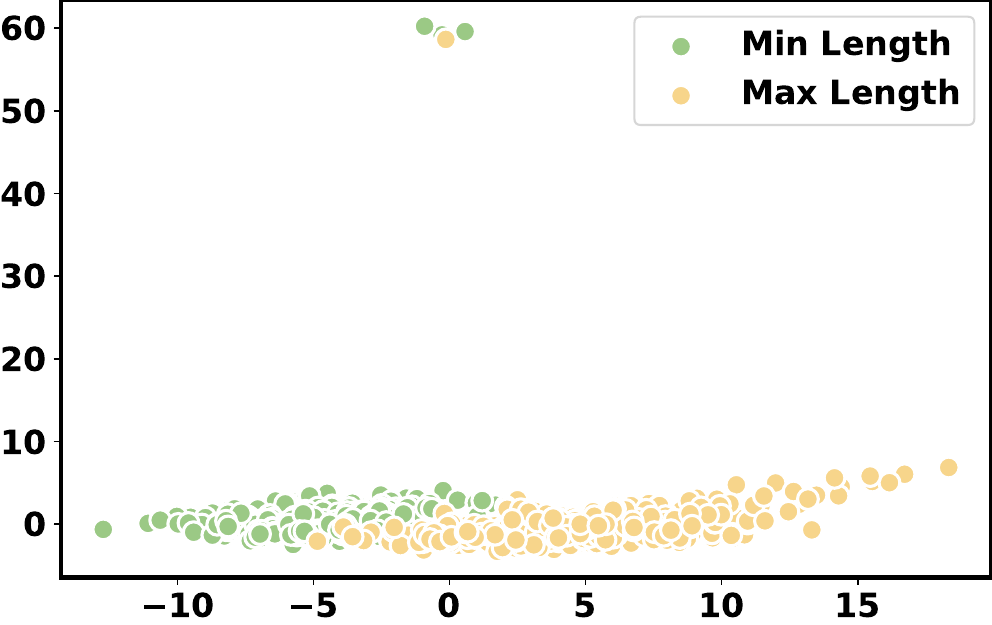}
    \caption*{(f) Layer 15 on MATH Level 3.}
  \end{minipage}
  \caption{Visualization of representational differences between the shortest and longest correct reasoning paths in R1-Distill-Qwen-7B.
Subfigures (a)–(c) illustrate the representation at layer 25 for math problems of increasing difficulty levels (Level 1, 3, and 5) from the MATH dataset. Subfigures (d)–(f) show the representations at different layers (layer 5, 10, and 15) for math problems at the same difficulty level (Level 3).}
  \label{fig:rep_analysis}
\end{figure*}

\subsection{Representation-level Dynamics}
\label{subsec:repre_analysis}

Building on the results presented in \S\ref{sec:lrm_efficiency}, we further explore the internal representational dynamics of LRMs by comparing the shortest and longest correct reasoning paths. We focus on the \texttt{R1-Distill-Qwen-7B} model for this visualization, with results for other models exhibiting the same trend. Specifically, we extract the hidden representations corresponding to the final reasoning step and apply Principal Component Analysis (PCA) to project them into two dimensions. The visualizations, shown in Figure~\ref{fig:rep_analysis}, highlight two key findings:

First, \textbf{across different problem difficulty levels, there is a pronounced shift between the shortest and longest reasoning paths in the representation space.} As illustrated in Figure \ref{fig:rep_analysis} (a)–(c), this separation remains consistent across MATH Level 1, Level 3, and Level 5, suggesting that efficient and verbose reasoning paths are systematically encoded in different regions of the hidden space regardless of task complexity.

Second, \textbf{this representational shift is consistently present across different layers of the model.} As shown in Figure \ref{fig:rep_analysis} (d)–(f), we observe similar separation patterns at layer 5, 10, and 15 on MATH Level 3 problems, indicating that the contrast between efficient and verbose reasoning paths is distributed throughout the model's different layers rather than confined to specific stages of reasoning.

This systematic shift in representational space suggests that LRMs encode an implicit directionality associated with reasoning efficiency. In the following sections, we examine whether this direction can be explicitly extracted and leveraged to steer LRMs toward more efficient reasoning behavior via targeted, interpretable, and training-free interventions.

\begin{wrapfigure}{r}{0.7\textwidth}
\centering
\small
\vspace{-4mm}
\caption{Lexical analysis with different LRMs on MATH Level 1 and Level 5.}
\label{tab:keyword}
\begin{tabular}{lcc|cc|cc}
\toprule
& \multicolumn{2}{c|}{\textbf{\textit{Wait}}} & \multicolumn{2}{c|}{\textbf{\textit{Alternatively}}} & \multicolumn{2}{c}{\textbf{\textit{However}}} \\
& \textbf{Min} & \textbf{Max} & \textbf{Min} & \textbf{Max} & \textbf{Min} & \textbf{Max} \\
\midrule
\rowcolor{gray!20} \multicolumn{7}{c}{\textbf{MATH Level 1}} \\
R1-Distill-Qwen-1.5B &3.04 &18.04 &0.88 &5.49 &0.01 &0.10 \\
R1-Distill-Qwen-7B &2.49 &9.69 &0.81 &3.75 &0.01 &0.11 \\
R1-Distill-Qwen-14B &2.35 &9.04 &0.62 &3.25 &0.02 &0.09 \\
GLM-Z1-9B &2.48 &6.35 &0.82 &2.77 &0.08 &0.44 \\
GLM-Z1-32B &2.38 &5.55 &0.71 &2.57 &0.09 &0.30 \\
\midrule
\midrule
\rowcolor{gray!20} \multicolumn{7}{c}{\textbf{MATH Level 5}} \\
R1-Distill-Qwen-1.5B &15.43 &52.35 &2.47 &6.02 &0.12 &0.25 \\
R1-Distill-Qwen-7B &10.89 &33.71 &2.26 &5.49 &0.16 &0.34 \\
R1-Distill-Qwen-14B &11.26 &30.94 &2.26 &5.27 &0.17 &0.36 \\
GLM-Z1-9B &10.37 &24.04 &3.09 &6.32 &1.44 &4.29 \\
GLM-Z1-32B &12.92 &16.82 &3.91 &5.09 &1.82 &2.53 \\
\bottomrule
\end{tabular}
\end{wrapfigure}

\subsection{Behavioral-Level Correlates}
\label{subsec:lex_analysis}

We further examine how the representation shift influences the content and reasoning structure of model outputs. Prior research suggests that inefficiency in LRMs primarily stems from frequent self-reflection, where models continue to explore alternative reasoning paths even after having arrived at the correct answer \citep{chen2024not,zhang2025reasoning}. This tendency can be quantitatively measured by lexical and reasoning behavior indicators.

For lexical analysis, we follow existing work \citep{guo2025deepseek,zeng2025simplerl,yeo2025demystifying} and compute the frequency of indicative self-reflection keywords---such as ``wait'', ``alternatively'', and ``however''---in both the shortest and longest correct reasoning paths. In Table \ref{tab:keyword}, we observe a consistent pattern across various LRMs: shorter reasoning paths exhibit significantly fewer instances of such keywords, suggesting a lower degree of self-reflective behavior and, consequently, improved reasoning efficiency.

\begin{wrapfigure}{r}{0.65\textwidth}
  \centering
  \includegraphics[width=1.0\linewidth]{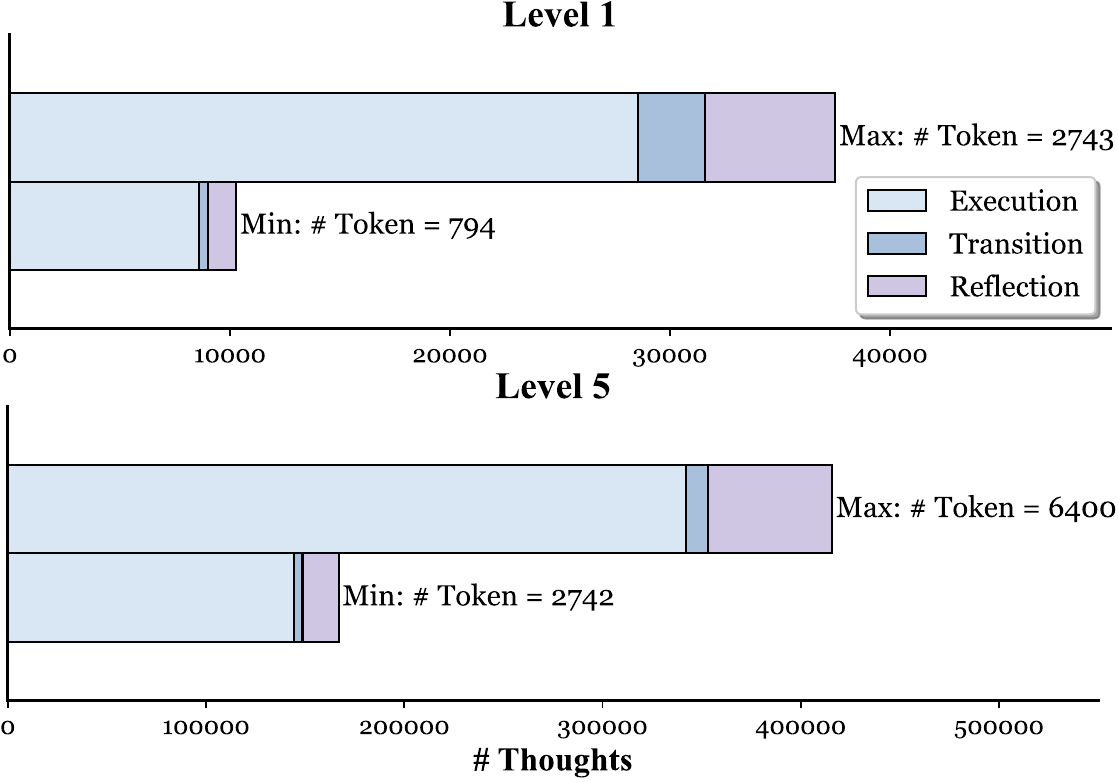}
  \vspace{-2mm}
  \caption{Visualization of reasoning behavior distributions across reasoning paths of varying lengths on MATH Level 1 and Level 5. The backbone is R1-Distill-Qwen-7B.}
  \label{fig:qwq_behavior}
  \vspace{-4mm}
\end{wrapfigure}For reasoning behavior analysis, we adopt the framework proposed by \citet{chen2025seal}, which segments the model's thought process into three functional categories: \textit{Execution}, \textit{Reflection}, and \textit{Transition}. Among these, the \textit{Reflection} and \textit{Transition} phases are particularly indicative of self-reflective tendencies. Results in Figure \ref{fig:qwq_behavior} shows that shorter reasoning paths consistently involve fewer \textit{Reflection} and \textit{Transition} segments, further supporting the claim that efficient reasoning is associated with reduced self-reflection overhead. Detailed definitions of the three reasoning stages are provided in Appendix \ref{app:behavior}.

\section{Efficiency Steering}
\label{sec:eff_direction}

Given the observation in \S\ref{subsec:repre_analysis} that efficient and verbose reasoning paths exhibit clear shifts in the representational space of LRMs, a natural question arises: can we directly manipulate the internal representations of LRMs to promote and facilitate efficient reasoning? In this section, we provide a definitive answer to this research question.

Specifically, we first provide the background on the structure of the representation space in transformer-based LRMs (\S\ref{subsec:pre}). We then demonstrate that the reasoning efficiency of LRMs can be controllably modulated along a single direction in their representational space (\S\ref{subsec:eff_control}). Finally, we show that this method can be applied across a variety of existing LRMs, enabling a training-free approach to enhancing reasoning efficiency without compromising overall performance (\S\ref{subsec:eff_steering}).

\subsection{Preliminaries}
\label{subsec:pre}

\paragraph{Transformers} In decoder-only transformers \citep{vaswani2017attention}, each input token $t_i$ is mapped to a hidden representation through a series of transformations across $L$ layers. At the core of this process is the \emph{residual stream}, denoted as $\boldsymbol{h}_i^{l} \in \mathbb{R}^{d_{\text{model}}}$, which encodes the evolving internal state of the model for token $i$ at layer $l$. The residual stream is initialized from the token embedding, $\boldsymbol{h}_i^{0} = \mathtt{Embed}(t_i)$, and then updated at each layer through attention and feedforward (MLP) blocks:
\begin{align}
\tilde{\boldsymbol{h}}_i^{l} = \boldsymbol{h}_i^{l} + \mathtt{Attn}^{l}(\boldsymbol{h}_{1:i}^{l}), \quad \boldsymbol{h}_i^{l+1} = \tilde{\boldsymbol{h}}_i^{l} + \mathtt{MLP}^{l}(\tilde{\boldsymbol{h}}_i^{l}).
\end{align}
In our study, we focus on the residual stream $\boldsymbol{h}_i^{l}$ as the primary carrier of semantic and reasoning information. By analyzing and intervening in these representations, particularly at the final reasoning token, we aim to modulate the model's reasoning behavior in a targeted and interpretable manner.

\paragraph{Linear Representation Hypothesis} The linear representation hypothesis suggests that many human-interpretable attributes, such as sentiment, formality, or factuality, are encoded in approximately linear subspaces within the residual stream of language models \citep{mikolov2013linguistic,nanda2023emergent,zou2023representation,park2024linear}. Building upon this idea, recent work has explored controlling model behavior by manipulating hidden representations at inference time, without retraining. For example, linear directions have been used to make models more truthful \citep{li2023inference,campbell2023localizing,zhang2024truthx} and more harmless \citep{lee2024mechanistic,uppaal2024detox,zhao2025adasteer}. These successes demonstrate that interpretable behavioral properties can often be isolated and modified through simple linear operations in the representation space.

Motivated by these findings, we hypothesize that such inherent reasoning efficiency within LRMs may also be mediated by a similar structure, namely, a single direction that distinguishes efficient from verbose reasoning behavior.

\subsection{Efficiency Control in LRMs}
\label{subsec:eff_control}

To investigate whether reasoning efficiency can be explicitly controlled, we identify and manipulate a latent ``efficiency direction'' in LRMs using the \emph{difference-in-means} method \citep{belrose2023diff}. Specifically, we collect the hidden representations corresponding to the final reasoning token from both the shortest and longest correct reasoning paths. By computing the mean difference between these two sets of activations, we derive a single vector that captures the contrast between efficient and verbose reasoning behavior.

\paragraph{Difference-in-means.}
To extract the efficiency direction from the model's residual stream, we compute the difference between average hidden activations for efficient and verbose reasoning samples. This technique, known as \emph{difference-in-means}, has proven effective in isolating behaviorally meaningful directions in prior work \citep{marks2023geometry,tigges2023linear,arditi2024refusal}. For each transformer layer $l \in [L]$, we compute the mean representation at the final token position for both sets:
\begin{align}
\boldsymbol{\mu}^{l}_{\text{\scriptsize efficient}} &= \frac{1}{|\mathcal{D}_{\text{efficient}}|} \sum\nolimits_{\mathbf{t} \in \mathcal{D}_{\text{efficient}}} \boldsymbol{h}^{l}(\mathbf{t}) , \quad  \\
\boldsymbol{\mu}^{l}_{\text{\scriptsize verbose}} &= \frac{1}{\lvert\mathcal{D}_{\text{verbose}}\rvert} \sum\nolimits_{\mathbf{t} \in \mathcal{D}_{\text{verbose}}} \boldsymbol{h}^{l}(\mathbf{t}).
\end{align}
where $\boldsymbol{h}^{l}(\mathbf{t})$ denotes the residual stream at the final reasoning token for input $\mathbf{t}$ at layer $l$.

We then define the efficiency direction at layer $l$ as:
\begin{equation}
\label{eq:diff_means}
\boldsymbol{v}^{l} = \boldsymbol{\mu}^{l}_{\text{\scriptsize efficient}} - \boldsymbol{\mu}^{l}_{\text{\scriptsize verbose}}
\end{equation}

Each direction $\boldsymbol{v}^{l}$ captures both the orientation along which efficient and verbose activations diverge, and the magnitude of their separation. This makes it a natural candidate for steering the model toward more efficient reasoning behavior.

\paragraph{Representation Intervention} Building on the linear representation hypothesis, we apply a lightweight intervention strategy that manipulates internal representations along behaviorally meaningful directions during inference. In our case, we steer the model's activations along the \emph{efficiency direction} $\boldsymbol{v}^l$ obtained via the difference-in-means method described above.

The intervention proceeds in two steps. First, we select a target layer $l$ and compute the direction vector $\boldsymbol{v}^l$ that separates efficient from verbose reasoning activations. Second, at inference time, we adjust the residual stream at the last token position of inputs in layer $l$ by injecting the direction vector, scaled by a steering coefficient $\lambda$:
\begin{equation}
\label{eq:steering}
\boldsymbol{h'}^{l} = \boldsymbol{h}^{l} + \lambda \, \boldsymbol{v}^l
\end{equation}
Here, $\boldsymbol{h}^{l}$ is the hidden state before intervention, and $\boldsymbol{h'}^{l}$ is the modified representation after steering. In our settings, we apply this intervention only at the final token of the input sequence.

\paragraph{Implementation Details} We use the GSM8K dataset \citep{cobbe2021gsm8k} to identify the efficiency direction. From this dataset, we sample 1,000 reasoning paths with the shortest lengths and 1,000 with the longest lengths, each selected among those that produce correct final answers. These samples form the efficient and verbose subsets used to compute the difference-in-means vectors. Our intervention experiments for efficiency control are implemented based on the \texttt{vLLM} framework \citep{kwon2023efficient} for efficient and scalable inference.

\begin{figure*}
  \centering
  \begin{minipage}[t]{0.49\textwidth}
    \centering
    \includegraphics[width=\linewidth]{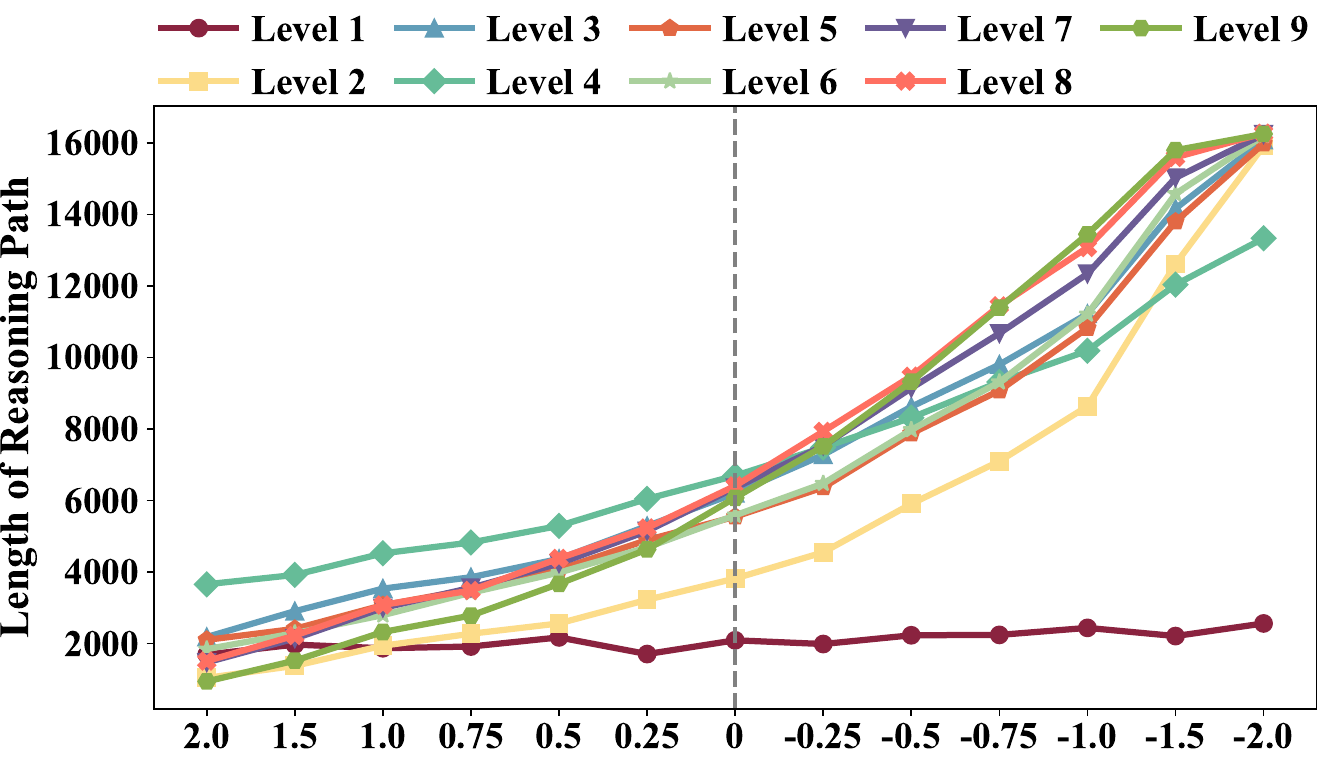}
    \caption*{(a) Impact of intervention strength on length.}
  \end{minipage}
  \begin{minipage}[t]{0.47\textwidth}
    \centering
    \includegraphics[width=\linewidth]{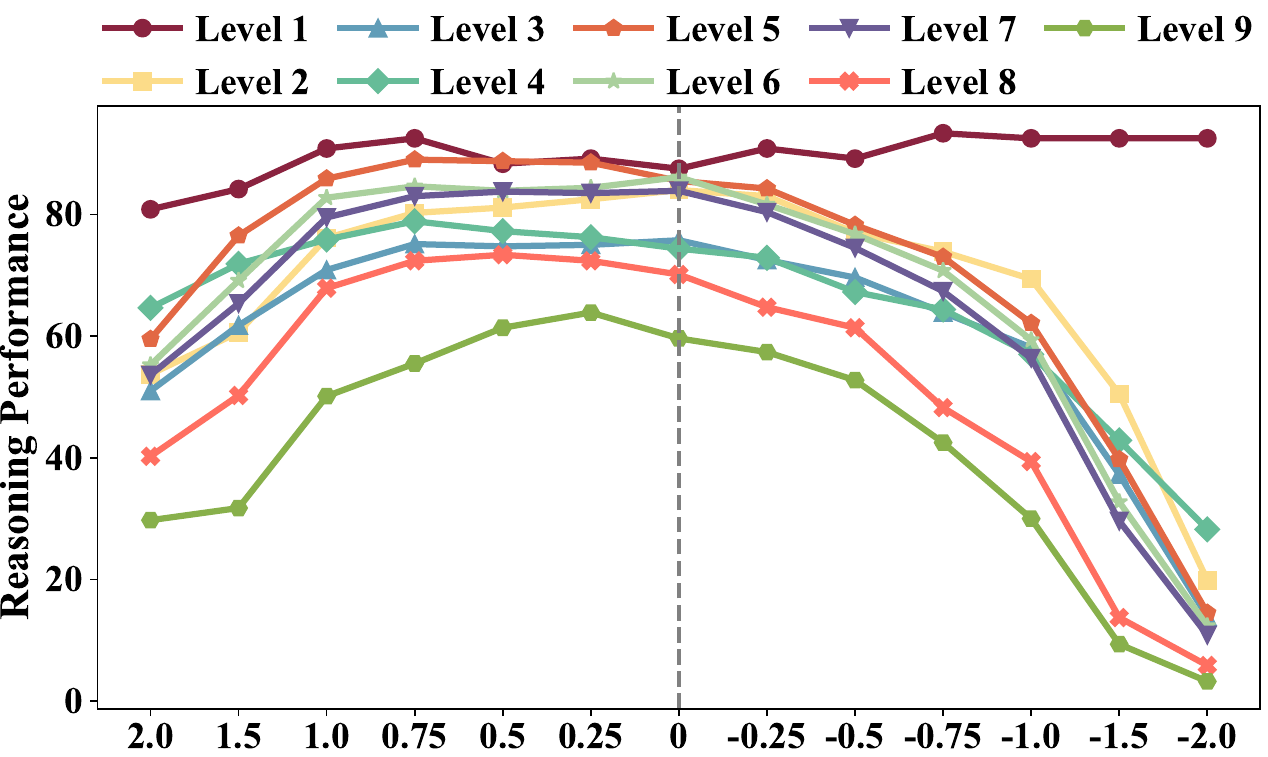}
    \caption*{(b) Impact of intervention strength on performance.}
  \end{minipage}
  \caption{Impact of intervention strength on (a) reasoning length and (b) performance across difficulty levels of R1-Distill-Qwen-7B on DeepMath dataset.}
  \label{fig:curve_strength}
\end{figure*}

\paragraph{Results and Analysis} To evaluate the impact of efficiency control, we conduct intervention experiments with the \texttt{R1-Distill-Qwen-7B} model on the DeepMath dataset \citep{he2025deepmath}, which consists of mathematical problems spanning nine difficulty levels. We choose DeepMath as the evaluation benchmark because it spans a wide range of difficulty levels, providing a comprehensive testbed for assessing the intervention's impact. Moreover, none of its samples overlap with the GSM8K dataset used to extract the efficiency direction, ensuring the generalizability of our conclusions.

Figure~\ref{fig:curve_strength} illustrates the relationship between the strength of this intervention---quantified by the steering coefficient $\lambda$---and the average length of the model's reasoning paths. Across all difficulty levels, we observe two clear and consistent trends:

\textbf{Reasoning efficiency in LRMs is controllable.} As $\lambda$ increases in the direction of efficiency, the model's reasoning becomes progressively shorter; conversely, steering in the opposite direction leads to longer and more redundant reasoning.

\textbf{Efficiency can be improved without compromising performance.} Within a reasonable range of $\lambda$, the model maintains its accuracy while achieving substantial reductions in reasoning length, indicating that efficient reasoning behavior can be induced without sacrificing correctness.

These results provide strong empirical evidence that a single, interpretable direction in the representation space can reliably mediate reasoning efficiency in LRMs. Crucially, this is achieved through a lightweight and training-free intervention, making it broadly accessible and scalable for future LRM deployment and optimization.

\subsection{Experimental Results}
\label{subsec:eff_steering}

\paragraph{Models} We evaluate the generality of our efficiency steering method across a diverse set of large reasoning models, consistent with the model list in Table~\ref{tab:model-overview} from Section~\ref{sec:lrm_efficiency}. In total, we include 7 LRMs covering different training paradigms, such as distilled models (e.g., DeepSeek-Qwen-Distill) and RL-optimized models (e.g., GLM-Z1-9B and QwQ-32B). This diverse selection ensures that our analysis captures a wide spectrum of reasoning behaviors and architectural scales.

\paragraph{Evaluation Configurations} We follow existing works to include the following evaluation datasets: MATH-500 \citep{lightman2023let}, AMC23 \citep{maa_amc}, and AIME 2024 and 2025 \citep{maa_amc}. Following the DeepSeek R1 configuration \citep{guo2025deepseek}, we set the maximum generation length—including both the reasoning trace and final answer—to 32,768 tokens for all models. For each test question, we sample 8 outputs using a temperature of 0.6 and a top-$p$ value of 0.95.

We report two main metrics: (1) \textbf{Performance}, measured by \textbf{pass@1} accuracy,
\[
\text{pass@1} = \frac{1}{k} \sum\nolimits_{i=1}^{k} p_i,
\]
where $k$ is the number of sampled outputs and $p_i$ denotes the correctness of the $i$-th response. This method provides reliable estimates of model accuracy across multiple samples. (2) \textbf{Length}, calculated as the average number of tokens (including both intermediate reasoning and the final answer) across all outputs on each test set. This metric reflects the reasoning efficiency of the model and is used to evaluate the effect of steering interventions.

\paragraph{Implementation Details} We use the same steering vector derived in \S\ref{subsec:eff_control} via the difference-in-means method. The intervention is consistently applied to the hidden state of the last input token only, following the procedure described in Equation~\ref{eq:steering}. This ensures a lightweight and targeted manipulation without altering the full sequence. For each model, the optimal steering coefficient $\lambda$ is selected empirically.

\begin{table*}
\centering
\setlength{\extrarowheight}{0pt}
\caption{\textbf{Evaluation of Efficiency Steering across four mathematical reasoning benchmarks.} We report both task accuracy (Performance ↑) and average reasoning trace length (Length ↓) for each model, before and after applying our method. Results show that Efficiency Steering consistently reduces reasoning length across all models and datasets, with minimal or no drop in performance—and in many cases, accuracy improves. This demonstrates the effectiveness of our training-free activation steering approach in inducing efficient reasoning behaviors without compromising correctness.}
\label{tab:main_exp}
\resizebox{\linewidth}{!}{
\begin{tabular}{lcc|cc|cc|cc}
\toprule
\textbf{}  & \multicolumn{2}{c|}{\textbf{MATH-500}} & \multicolumn{2}{c|}{\textbf{AMC}} & \multicolumn{2}{c|}{\textbf{AIME 2024}} &\multicolumn{2}{c}{\textbf{AIME 2025}} \\
& \textbf{Performance $\uparrow$} & \textbf{Length $\downarrow$} & \textbf{Performance $\uparrow$} & \textbf{Length $\downarrow$} & \textbf{Performance $\uparrow$} & \textbf{Length $\downarrow$} & \textbf{Performance $\uparrow$} & \textbf{Length $\downarrow$} \\
\midrule
R1-Distill-Qwen-1.5B &83.40 &4317.08 &58.73 &8251.63 &27.08 &11886.86 &25.83 &10700.73 \\
+ Efficiency. Steering &\textbf{84.60} &\textbf{3899.99} &\textbf{62.80} &\textbf{7510.97} &\textbf{27.92} &\textbf{11135.50} &\textbf{28.33} &\textbf{9746.06} \\
\midrule
R1-Distill-Qwen-7B &92.20 &3495.61 &79.97 &6357.50 &49.17 &10199.45 &35.00 &10518.44 \\
+ Efficiency. Steering &\textbf{92.60} &\textbf{2559.78} &\textbf{80.27} &\textbf{4829.29} &\textbf{50.00} &\textbf{7827.73} &\textbf{41.67} &\textbf{7982.68} \\
\midrule
R1-Distill-Qwen-14B &93.40 &3279.73 &85.24 &5750.04 &65.00 &8931.70 &42.50 &10344.77 \\
+ Efficiency. Steering &92.20 &\textbf{2505.18} &\textbf{86.14} &\textbf{4334.03} &\textbf{65.83} &\textbf{5624.14} &\textbf{44.17} &\textbf{9881.04}\\
\midrule
R1-Distill-LLaMA-8B &89.40 &3820.06 &78.01 &6920.97 &45.00 &11757.59 &31.67 &11158.48 \\
+ Efficiency. Steering &88.00 &\textbf{3077.36} &\textbf{78.31} &\textbf{6063.62} &\textbf{46.67} &\textbf{10324.59} &\textbf{40.00} &\textbf{9888.52} \\
\midrule
\midrule
GLM-Z1-9B &95.80 &3115.67 &89.46 &6013.77 &72.08 &10055.56 &49.17 &11804.96 \\
+ Efficiency. Steering &\textbf{96.60} &\textbf{2403.60} &\textbf{90.21} &\textbf{5271.93} &\textbf{74.17} &\textbf{9187.34} &\textbf{55.00} &\textbf{10961.20} \\
\midrule
GLM-Z1-32B &96.00 &2872.75 &90.36 &5701.26 &75.00 &9118.65 &55.83 &10432.54 \\
+ Efficiency. Steering &\textbf{96.20} &\textbf{2207.53} &\textbf{91.72} &\textbf{5218.15} &73.33 &\textbf{8293.39} &\textbf{60.83} &\textbf{9819.23} \\
\midrule
QwQ-32B &96.40 &3885.94 &94.43 &8360.41 &78.75 &13385.95 &64.17 &15608.45 \\
+ Efficiency. Steering &96.20 &\textbf{3161.19} &\textbf{95.18} &\textbf{7922.91} &\textbf{80.42} &\textbf{12618.16} &\textbf{65.83} &\textbf{14047.87} \\
\bottomrule
\end{tabular}
}
\end{table*}

\paragraph{Results and Analysis} Table~\ref{tab:main_exp} summarizes the performance and reasoning length across four benchmarks and nine LRMs, both with and without efficiency steering. From the results, we can draw two key conclusions:

\textbf{Efficiency Steering consistently reduces reasoning overhead.} Across all 7 models and 4 benchmarks, steering in the direction of efficiency leads to a significant reduction in the average length of generated reasoning traces. Notably, this improvement holds for models of varying architectures and training paradigms. For instance, in MATH-500, the reasoning length of R1-Distill-Qwen-7B drops from 3,495.61 to 2,559.78 tokens, and QwQ-32B decreases from 3,885.94 to 3,161.19 tokens. These consistent trends across datasets and model families demonstrate the generality and robustness of our steering method in promoting more efficient reasoning behavior.

\textbf{Efficiency Steering achieves these reductions without degrading—and often slightly improving—model accuracy.} One major concern with shortening reasoning paths is potential degradation in task performance. However, our results show that efficiency steering does not harm accuracy; in fact, it often slightly improves it. For instance, GLM-Z1-32B maintains its accuracy at 96.20 on MATH-500 while reducing average reasoning length by over 600 tokens. Similarly, R1-Distill-Qwen-14B shows an increase from 85.24 to 86.14 on AMC23 and from 44.17 to 44.57 on AIME 2025. These results confirm that our approach promotes not only brevity but also focus—helping models avoid unnecessary detours and reach the correct answer more directly.

\section{Self-Rewarded Efficiency RL}
\subsection{Reward Design}
While our representational analysis reveals a direction along which reasoning efficiency can be explicitly steered, our behavioral findings suggest that verbose and efficient reasoning trajectories also differ systematically in their linguistic and phase-level characteristics. Motivated by this, we propose a reinforcement learning framework---Self-Rewarded Efficiency RL---to explicitly reward efficient reasoning behaviors.

The central idea is to use model-generated rollouts as the basis for constructing dynamic, instance-specific rewards that encourage both accuracy and brevity. For each input $x_i$, we generate a set of responses $Y(x_i) = \{y_1, y_2, \dots, y_n\}$. A response $y_j$ receives a reward defined as:
\begin{equation}\label{eq:sol-reward}
r(y_j) = \lambda_1 \cdot \mathbb{I}(y_j = y_i^*) - \lambda_2 \cdot \text{max} (0, \ \ell(y_j) - \ell^{\text{Min\_Correct}}),
\end{equation}
where $\mathbb{I}(y_j = y_i^*)$ denotes whether the response matches the correct answer, and $\ell(y_j)$ is the number of tokens in the reasoning trace. The minimal length among all correct responses in the current rollout is computed as:
\begin{equation}\label{eq:sol-reward}
\ell^{\text{Min\_Correct}} = \min_{y_j \in Y(x_i):\,\mathbb{I}(y_j = y_i^*)=1} \ell(y_j)
\end{equation}
This reward function encourages the model to produce shorter correct answers by penalizing deviations from the shortest correct reasoning path found in the current generation batch. When no correct answer exists, the average length serves as a soft regularizer to discourage uniformly verbose outputs.

Compared to conventional reward design—which often only considers correctness—our formulation introduces instance-adaptive efficiency pressure based on the model’s own generation history. This aligns with our empirical observation that, even within the same model, shorter reasoning paths with equal correctness do exist but are underutilized. By guiding optimization toward such paths, Self-Rewarded Efficiency RL serves as a behaviorally grounded counterpart to Efficiency Steering, providing a learning-based alternative for activating latent reasoning efficiency in LRMs.

\subsection{Policy Optimization via GRPO}
\label{subsec:grpo}
To optimize the model under the self-rewarded efficiency objective, we adopt the Group Relative Policy Optimization (GRPO) \citep{shao2024deepseekmath}, a variant of policy gradient methods tailored for stabilizing optimization. More specifically, for each question $q$, GRPO samples a group of outputs $\{o_1, o_2, \cdots, o_G\}$  from the old policy  $\pi_{\theta_{old}}$  and then optimizes the policy model by maximizing the following objective:
\begin{equation}\scriptsize
\begin{split}
    \mathcal{J}_{\text{GRPO}}(\theta) &= \mathbb{E}{[q \sim P(Q), \{o_i\}_{i=1}^G \sim \pi_{\theta_{\text{old}}}(O|q)]}  \\
    & \frac{1}{G}\sum_{i=1}^G\frac{1}{|o_i|} \sum_{t=1}^{|o_i|} \left\{ \min \left[ \frac{\pi_\theta(o_{i,t} | q, o_{i,<t})}{\pi_{\theta_{old}}(o_{i,t} | q, o_{i,<t})} \hat{A}_{i,t}, \text{clip} \left( \frac{\pi_\theta(o_{i,t} | q, o_{i,<t})}{\pi_{\theta_{\text{old}}}(o_{i,t} | q, o_{i,<t})}, 1 - \epsilon, 1 + \epsilon \right)  \hat{A}_{i,t} \right] - \beta \mathbb{D}_{KL}\left[\pi_{\theta} || \pi_{\text{ref}}\right]\right\} ,
\end{split}
\label{eq:GRPO-obj}
\end{equation}
where $\epsilon$ and $\beta$ are hyper-parameters, and $\hat{A}_{i,t}$ is the advantage calculated based on relative rewards of the outputs inside each group only, i.e., $\hat{A}_{i, t} = \widetilde{r}_i = \frac{r_i- {\rm mean}(\mathbf{r})}{{\rm std}(\mathbf{r})}$.

\subsection{Implementation Details}

We implement our Self-Rewarded Efficiency RL framework using the \texttt{OpenRLHF} library \citep{hu2024openrlhf}, and apply the GRPO across models ranging from 1.5B to 14B parameters. For training the 14B model, we use 16 A800 GPUs, while all other model sizes are trained on 8 A800 GPUs. Each training run consists of a single episode, with 8 rollouts per input sample.

Despite the limited episode count, we observe that model performance is highly sensitive to optimization steps beyond 30, where efficiency improves but correctness starts to deteriorate. We analyze this trade-off phenomenon in detail in \S\ref{subsec:duration}. To improve reasoning efficiency, we conduct training on Level 1–3 difficulty data from the MATH dataset, which empirically yields more favorable results compared to training on harder samples; further analysis is provided in \S\ref{subsec:training-data}.

We train for 1 epoch using a batch size of 16, with the maximum input length set to 1,024 tokens and maximum output length set to 8,192 tokens to accommodate long-form reasoning traces. The actor model is optimized using the Adam optimizer with a learning rate of $1 \times 10^{-6}$, and a KL penalty coefficient of 0.01 is applied to prevent policy drift from the initial actor. The reward weights are set as $\lambda_1 = 1$ and $\lambda_2 = 0.001$, balancing correctness with a mild penalty on deviation from the shortest correct path per rollout.

\begin{table*}
\centering
\setlength{\extrarowheight}{0pt}
\caption{\textbf{Evaluation of Self-Rewarded Efficiency RL across four mathematical reasoning benchmarks.} We report both task accuracy (Performance ↑) and average reasoning trace length (Length ↓) for each model, before and after applying our method. Results show that Self-Rewarded Efficiency RL consistently reduces reasoning length across all models and datasets, with minimal or no drop in performance—and in many cases, accuracy improves. This demonstrates the effectiveness of our reinforcement approach in inducing efficient reasoning behaviors without compromising correctness.}
\label{tab:main_exp_rl}
\resizebox{\linewidth}{!}{
\begin{tabular}{lcc|cc|cc|cc}
\toprule
\textbf{}  & \multicolumn{2}{c|}{\textbf{MATH-500}} & \multicolumn{2}{c|}{\textbf{AMC}} & \multicolumn{2}{c|}{\textbf{AIME 2024}} &\multicolumn{2}{c}{\textbf{AIME 2025}} \\
& \textbf{Performance $\uparrow$} & \textbf{Length $\downarrow$} & \textbf{Performance $\uparrow$} & \textbf{Length $\downarrow$} & \textbf{Performance $\uparrow$} & \textbf{Length $\downarrow$} & \textbf{Performance $\uparrow$} & \textbf{Length $\downarrow$} \\
\midrule
R1-Distill-Qwen-1.5B &\textbf{83.40} &4317.08 &58.73 &8251.63 &27.08 &11886.86 &\textbf{25.83} &10700.73 \\
+ S-R Efficiency RL &83.00 &\textbf{2400.18} &\textbf{62.65} &\textbf{4970.46} &\textbf{28.33} &\textbf{9024.70} &24.17 &\textbf{7458.68} \\
\midrule
R1-Distill-Qwen-7B &\textbf{92.20} &3495.61 &\textbf{79.97} &6357.50 &49.17 &10199.45 &35.00 &10518.44 \\
+ S-R Efficiency RL &86.80 &\textbf{1168.03} &79.07 &\textbf{3189.12} &\textbf{51.67} &\textbf{6581.56} &\textbf{40.83} &\textbf{6405.04} \\
\midrule
R1-Distill-Qwen-14B &\textbf{93.40} &3279.73 &85.24 &5750.04 &65.00 &8931.70 &\textbf{42.50} &10344.77 \\
+ S-R Efficiency RL &92.40 &\textbf{1810.67} &\textbf{85.69} &\textbf{3568.09} &\textbf{66.25} &\textbf{6972.27} &41.67 &\textbf{7279.68} \\
\midrule
\midrule
GLM-Z1-9B &95.80 &3115.67 &\textbf{89.46} &6013.77 &72.08 &10055.56 &49.17 &11804.96 \\
+ S-R Efficiency RL &\textbf{96.40} &\textbf{1473.89} &89.01 &\textbf{4010.12} &\textbf{72.92} &\textbf{8468.15} &\textbf{52.50} &\textbf{9001.35} \\
\bottomrule
\end{tabular}
}
\end{table*}

\subsection{Overall Results}

We evaluate the effectiveness of our proposed Self-Rewarded Efficiency RL (S-R Efficiency RL) across four mathematical reasoning benchmarks: MATH-500, AMC, AIME 2024, and AIME 2025. The results are summarized in Table \ref{tab:main_exp_rl}, where we report both Performance (i.e., task accuracy in \%) and Length (i.e., average token length of reasoning traces) for each model before and after applying our method.

\paragraph{Consistent Efficiency Gains Across All Models}
S-R Efficiency RL consistently reduces the average reasoning length by a large margin across all datasets and all model scales. For instance, on MATH-500, the average reasoning length of R1-Distill-Qwen-1.5B drops from 4,317.08 tokens to 2,400.18 (-44.4\%), while maintaining similar accuracy (from 83.40\% to 83.00\%). For larger models such as R1-Distill-Qwen-14B, the token length is reduced from 3,279.73 to 1,810.67 (-44.8\%), with only a marginal performance drop (-1.2\%). This trend is consistent across other datasets, demonstrating the robustness and scalability of our method in enforcing efficient reasoning.

\paragraph{No Performance Sacrifice—and Sometimes Improvement}
Notably, on many experimental settings, efficiency gains are achieved without performance degradation, and in several cases, accuracy even improves. For example, GLM-Z1-9B improves from 95.80\% to 96.40\% on MATH-500 and from 49.17\% to 52.50\% on AIME 2025, along with reasoning lengths reduced by 52.7\% and 23.7\% respectively. This illustrates that inducing concise reasoning does not inherently impair the model’s ability to reach correct conclusions, especially when guided by adaptive instance-level rewards.

\paragraph{Robustness Across Dataset Difficulty}
Across datasets of varying difficulty—from the easier AMC benchmark to the more challenging AIME 2025—the S-R Efficiency RL framework shows strong generalization. For instance, R1-Distill-Qwen-7B reduces its average reasoning length on AMC from 6,357.50 to 3,189.12, while also achieving a performance gain from 79.97\% to 81.63\%. Similarly, on the hardest benchmark (AIME 2025), the same model reduces length by 26.4\% and improves performance by over 4 points.

\paragraph{Benefits Scale with Model Size}
Larger models appear to benefit more from our efficiency reinforcement strategy. On AIME 2024, R1-Distill-Qwen-14B improves from 65.00\% to 66.25\% while shortening reasoning traces by over 22\% (from 8,931.70 to 6,972.27). This suggests that large models, which tend to overthink more, may contain greater latent efficiency potential that can be unlocked with our method.

\paragraph{Summary}
Overall, the results validate the effectiveness of S-R Efficiency RL in inducing concise and effective reasoning without sacrificing accuracy. These findings support our hypothesis that LRMs already possess inherent efficiency, which can be surfaced and reinforced through self-guided, reward-based reinforcement learning.

\begin{figure*}
  \centering
  \begin{minipage}[t]{0.47\textwidth}
    \centering
    \includegraphics[width=\linewidth]{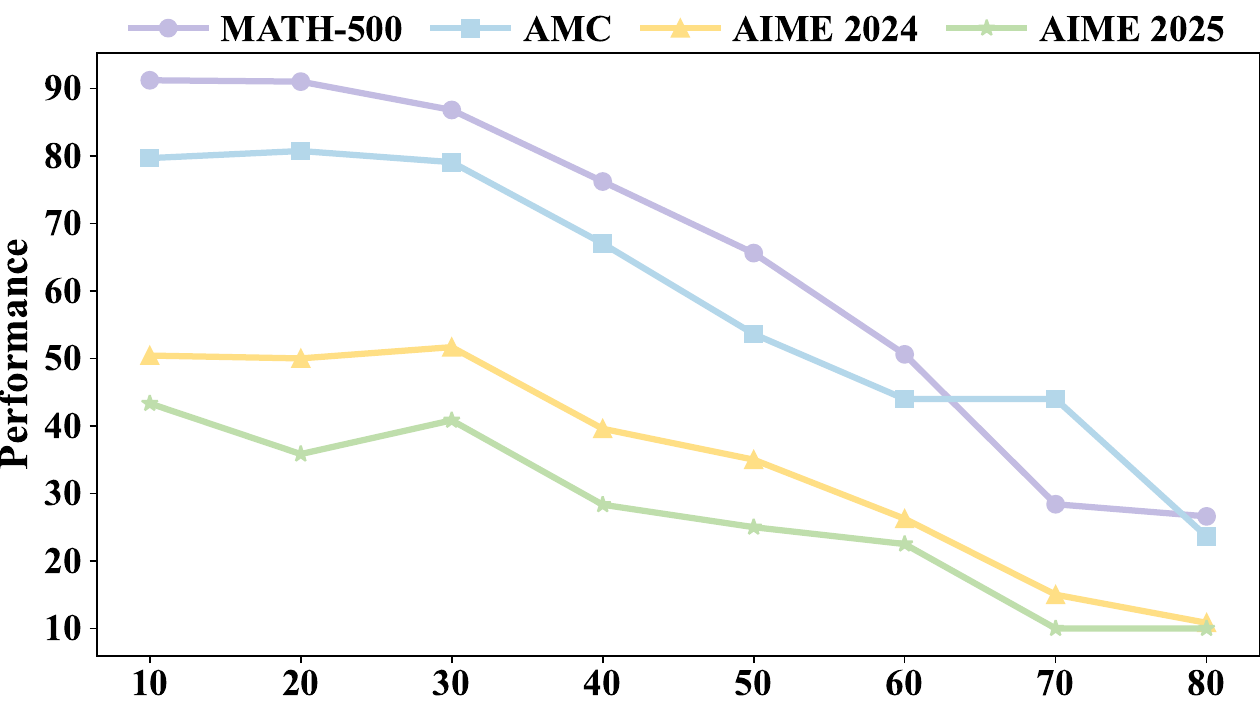}
    \caption*{(a) Accuracy over training steps.}
  \end{minipage}
  \begin{minipage}[t]{0.49\textwidth}
    \centering
    \includegraphics[width=\linewidth]{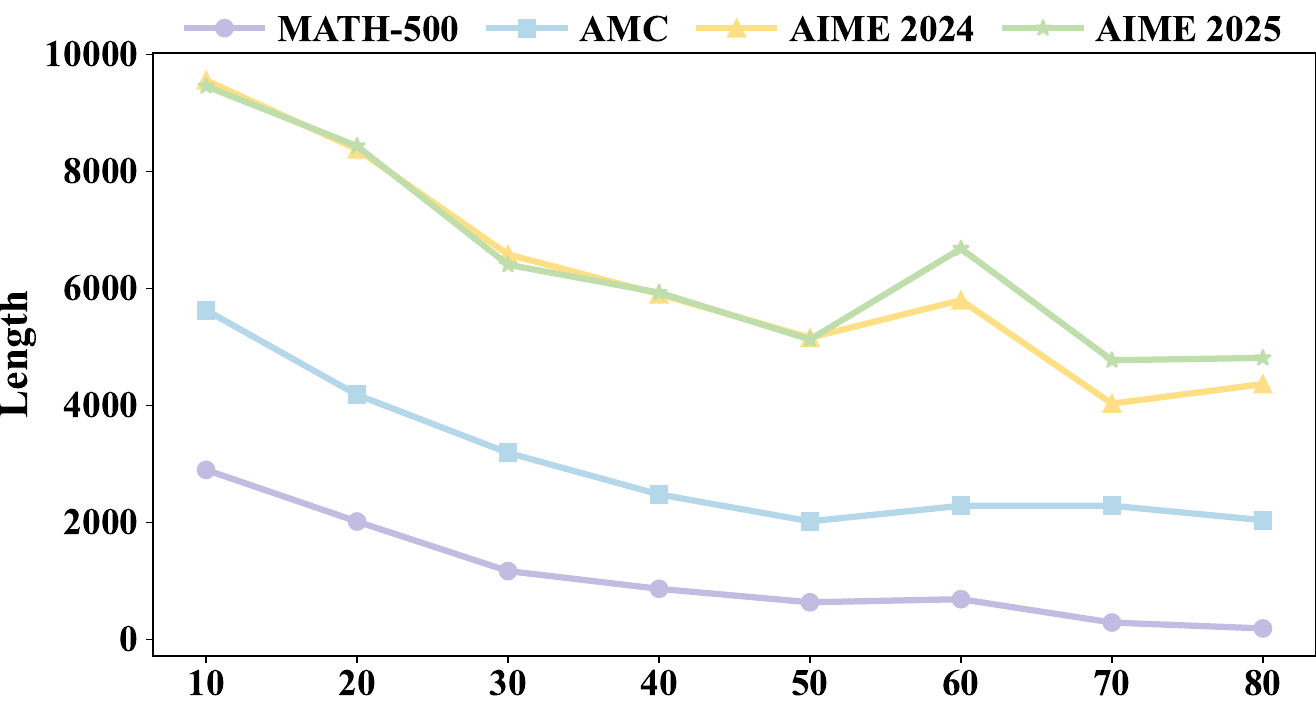}
    \caption*{(b) Average reasoning length over training epochs.}
  \end{minipage}
  \caption{\textbf{Training dynamics of S-R Efficiency RL on four reasoning benchmarks.} 
(a) Accuracy over training steps. (b) Average reasoning length over training epochs. 
Results are reported for R1-Distill-Qwen-7B. Significant efficiency improvements are achieved within the first 30 steps, with mild or no degradation in performance across most datasets.}
\label{fig:training-duration}
\end{figure*}

\subsection{Analysis on Training Duration}
\label{subsec:duration}

To further understand the dynamics of Self-Rewarded Efficiency RL, we analyze how model performance and reasoning efficiency evolve over the course of training. Figure~\ref{fig:training-duration} presents the results of fine-tuning R1-Distill-Qwen-7B with S-R Efficiency RL for 10 to 80 steps, evaluated on four benchmarks: MATH-500, AMC, AIME 2024, and AIME 2025.

\paragraph{Efficiency Improvements Emerge Early}
As shown in Figure~\ref{fig:training-duration}(b), significant reductions in average reasoning length occur within the first 10–30 steps. For instance, on MATH-500, the average length drops from over 3400 tokens to approximately 2200 tokens within 20 steps, eventually reaching below 1500 tokens after 80 steps. Similar early gains are observed on AMC and AIME datasets. This demonstrates that our reward design provides strong learning signals, enabling the model to quickly acquire efficient reasoning patterns.

\paragraph{Performance Remains Stable with Mild Trade-offs}
Figure~\ref{fig:training-duration}(a) shows that accuracy remains largely stable across most datasets, especially within the first 40 steps. On MATH-500 and AMC, performance drops are minimal (<2 points) despite substantial length reduction. On harder datasets like AIME 2024 and AIME 2025, a modest trade-off is observed beyond 40 steps, where further compression leads to a slight decline in accuracy. For example, on AIME 2025, accuracy decreases from ~39\% to ~32\% between 30 and 70 steps. This illustrates a natural trade-off frontier between brevity and correctness, especially on complex problems requiring more extensive multi-step reasoning.

\paragraph{Dataset Difficulty Affects Convergence Behavior}
The relative smoothness and monotonicity of the length curves vary across datasets. On simpler datasets like MATH-500 and AMC, both performance and length exhibit stable trends. In contrast, AIME 2025 shows more fluctuation in both metrics—particularly a spike in reasoning length around step 60—suggesting greater instability under tighter efficiency constraints. This highlights the need for adaptive or curriculum-aware scheduling when applying S-R Efficiency RL to more challenging reasoning domains.

\paragraph{Summary}
Overall, the analysis confirms that S-R Efficiency RL induces fast and stable efficiency gains, particularly in the early stages of training. While aggressive compression may introduce slight performance degradation on harder tasks, the trade-off is controllable and often worth the efficiency gain. Importantly, the fact that most efficiency improvements occur within the first 10–30 steps significantly reduces the required training duration, making the method computationally affordable. This property enhances the practicality of S-R Efficiency RL, especially in scenarios where inference cost, latency, or token budget are critical constraints.

\begin{figure*}
  \centering
  \begin{minipage}[t]{0.48\textwidth}
    \centering
    \includegraphics[width=0.9\linewidth]{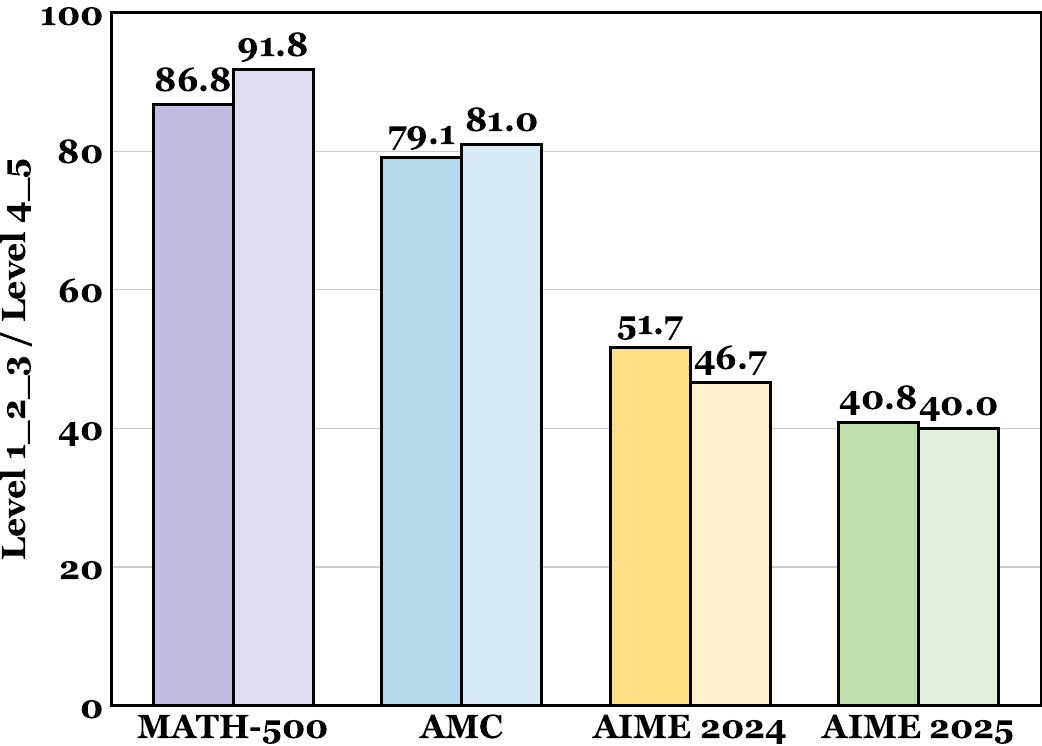}
    \caption*{(a) Effect of training data difficulty on reasoning performance.}
  \end{minipage}
  \begin{minipage}[t]{0.48\textwidth}
    \centering
    \includegraphics[width=0.9\linewidth]{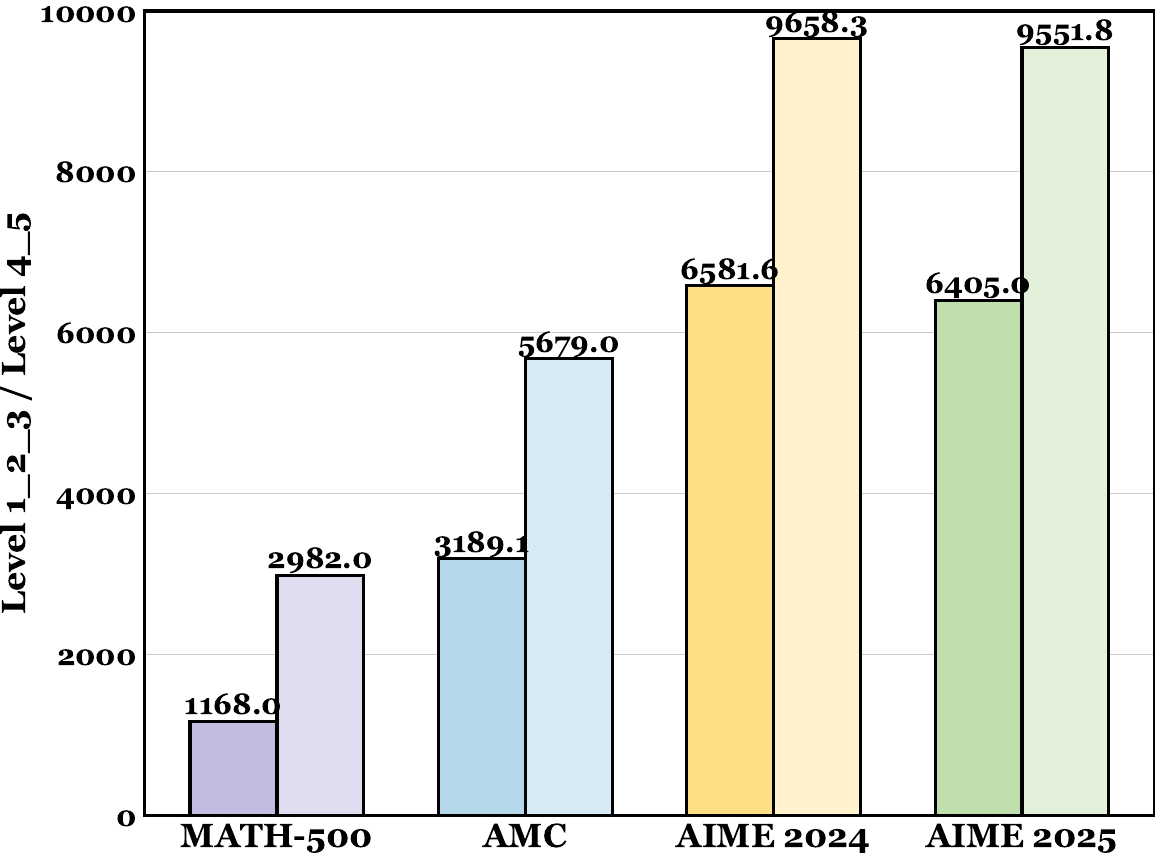}
    \caption*{(b) Effect of training data difficulty on reasoning efficiency.}
  \end{minipage}
  \caption{\textbf{Effect of training data difficulty on reasoning performance and efficiency.} 
We compare models trained with only Level 1–3 (easy) vs. Level 4–5 (hard) data on four benchmarks. 
(a) Accuracy remains comparable across training regimes. 
(b) Models trained on easier data consistently produce significantly shorter reasoning traces, indicating better efficiency generalization.}
\label{fig:training-data}
\end{figure*}

\subsection{Analysis on Training Data}
\label{subsec:training-data}

To better understand how training data difficulty affects the efficacy of Self-Rewarded Efficiency RL, we conduct an ablation experiment by partitioning the MATH dataset into two subsets: (1) Level 1\_2\_3: easier examples (difficulty levels 1–3), and (2) Level 4\_5: harder examples (levels 4–5).

We then train the model separately on each subset while evaluating both configurations on the full test sets of MATH-500, AMC, AIME 2024, and AIME 2025. The results are reported in Figure~\ref{fig:training-data}, with (a) showing task performance and (b) showing average reasoning length.

\paragraph{Training on Easier Data Yields Better Efficiency}
As shown in Figure~\ref{fig:training-data}(b), models trained on Level 1\_2\_3 data consistently generate shorter reasoning traces across all benchmarks. For example, on MATH-500, the model trained on easier data produces outputs averaging 1,168.0 tokens, compared to 2,982.0 tokens from the Level 4\_5-trained model—a 60.8\% reduction. Similar efficiency advantages are observed on AMC (-43.8\%) and AIME datasets (e.g., -32.9\% on AIME 2025). These results suggest that exposing the model to simpler, more direct reasoning trajectories during training facilitates more concise generation at test time.

\paragraph{Performance Remains Comparable Across Training Conditions}
Despite the difference in training data difficulty, the task accuracy of both models remains nearly identical across benchmarks, as shown in Figure~\ref{fig:training-data}(a). On MATH-500 and AMC, the gap is negligible, and even on harder tasks like AIME 2024 and AIME 2025, the difference is within 1 point. This highlights the strong generalization ability of models trained on easier problems, and indicates that efficiency-aware RL does not require difficult training data to generalize to difficult reasoning tasks.

\paragraph{Implications for Data Curation}
These findings have practical implications for training efficiency-aware LLMs. In resource-constrained settings, collecting or sampling easier problems for RL may provide better efficiency outcomes at lower training cost. It also points to the possibility that complex reasoning strategies can be effectively learned through self-improvement from simple examples—especially when guided by appropriate reward shaping mechanisms like ours.

\paragraph{Summary}
Training with simpler examples (Level 1\_2\_3) leads to more concise reasoning behavior while maintaining strong performance across both easy and hard benchmarks. This demonstrates that the efficiency prior learned from simpler data transfers well to more complex problems, underscoring the effectiveness and practicality of curriculum-aware or simplicity-biased training strategies in self-rewarded efficiency optimization.

\section{Related Works}

Large reasoning models have achieved remarkable reasoning performance by generating explicit chain-of-thought (CoT) solutions, but this often comes at the cost of reasoning inefficiency. Such verbose, redundant intermediate steps contribute to the overthinking phenomenon---improved accuracy via longer CoTs but with significant overhead in computation and latency \citep{chen2024not,wang2025harnessing,qu2025survey,feng2025efficient}. These observations highlight the urgent need for more efficient reasoning in LRMs, especially for multi-step mathematical or logical tasks.

\paragraph{Supervised Finetuning for Controlling Reasoning Length} A number of solutions have used supervised fine-tuning (SFT) to control the length or structure of model reasoning. One prominent direction is chain-of-thought compression. \citep{kang2025c3ot} propose C3oT, a framework that trains models to generate shorter reasoning traces without losing crucial information. C3oT introduces a dedicated ``compressor'' that transforms a long CoT into a concise version, and the model is fine-tuned on pairs of long and compressed rationales to learn to reason more succinctly. Another line of work uses distilled reasoning: \citep{munkhbat2025self} show that typical CoT outputs contain many redundant tokens, and they fine-tune LLMs on self-generated concise reasoning paths to eliminate needless steps. By training on the model's own best-of-$N$ sampled solutions filtered for brevity, their approach achieves about a 30\% reduction in tokens on GSM8K and MATH benchmarks with no drop in accuracy. Similarly, \citep{chen2024not} employ a self-training paradigm to mitigate overthinking, gathering streamlined reasoning examples and retraining the LLM to solve questions with fewer, more essential steps.

In addition to fine-tuning, clever prompt engineering can encourage efficient reasoning. For instance, simply instructing the model to ``think step-by-step concisely'' or to skip trivial sub-steps can cut down needless tokens. \citep{xu2025towards} guide models to adjust the level of detail based on predicted task difficulty, effectively telling the model to use shorter reasoning for easier problems. In extreme cases, a prompt that suppresses the chain-of-thought entirely (e.g., the NoThinking prompt) can yield substantial speed-ups: \citep{zhao2025trade} show that querying a model for the final answer directly, without any intermediate explanation, matched or surpassed CoT prompting on several tasks when constrained to the same total token limit. Overall, these supervised and prompt-based approaches attempt to reign in overthinking by either retraining models on compressed rationales or cleverly biasing their generated reasoning toward brevity.

\paragraph{Reinforcement Learning-Based Methods for Efficient Reasoning}

An alternative branch of work uses reinforcement learning (RL) to optimize reasoning trajectories for efficiency. Rather than relying on fixed training datasets, these methods define reward signals that penalize excessive reasoning or encourage timely correct answers. \citep{aggarwal2025l1} introduce L1, a model fine-tuned with a Length-Controlled Policy Optimization scheme to obey a target reasoning length given in its prompt. By optimizing for both accuracy and brevity, L1 can smoothly trade off computation for performance. Another representative work is ThinkPrune by \citep{hou2025thinkprune}, which uses RL to progressively prune long chains-of-thought. ThinkPrune trains the model with an added token budget: any generated reasoning beyond a set token limit yields zero reward, forcing the LLM to focus on the most critical steps. Notably, these methods integrate a notion of cost into the training objective (via length penalties or budgeted rewards), directly addressing the overthinking issue.

Finally, it is important to contrast these prior approaches with our proposed method. Most existing solutions rely on external feedback to instill efficiency. SFT-based approaches require curated short-CoT data (e.g. compressed rationales or distilled traces). RL-based methods, while powerful, hinge on pre-defined heuristics – for instance, a fixed token budget threshold in ThinkPrune. In contrast, our methods, both Efficiency Steering and Self-Rewarded Efficiency RL, do not require any external reward model or heuristic. It leverages the model's own internal evaluations as a form of self-reward, dynamically guiding the reasoning process without additional supervision. In other words, our approach allows the LRM to self-regulate its ``thinking'' in real time – shortening or simplifying its chain-of-thought when appropriate. This self-guided strategy stands apart from prior work by offering a flexible, on-the-fly efficiency enhancement that is adaptive and entirely free of external labels or costly optimization processes. Although concurrent work \citep{yi2025shorterbetter} also investigates self-reward mechanisms for length control, their study is confined to a single 7B model and does not examine key variables such as training duration or data distribution. In contrast, our work provides a more comprehensive evaluation across multiple model scales and conditions, offering deeper insights into the robustness and generalizability of self-guided efficiency optimization.

\section{Conclusion}
This work investigates the problem of overthinking in large reasoning models and reveals that reasoning inefficiency is not an inevitable artifact of LRM generation, but rather a controllable behavior embedded in the model’s internal representations. Through comprehensive analysis, we uncover that efficient reasoning traces are linearly separable in the representation space and exhibit consistent behavioral and lexical distinctions from verbose ones.
Motivated by these insights, we propose two efficiency enhancement strategies: Efficiency Steering, which introduces a training-free representational intervention vector to steer reasoning length during inference, and Self-Rewarded Efficiency RL, which applies dynamic reward shaping to optimize for correctness and brevity jointly. Our methods are model-agnostic, scalable, and require minimal overhead.
Empirical results on a suite of LRM backbones and reasoning tasks show that both methods significantly reduce reasoning length without sacrificing accuracy. These findings underscore that reasoning efficiency can be surfaced and enhanced through simple yet effective mechanisms that align with the model’s intrinsic structure, offering new directions for developing more cost-efficient and user-friendly reasoning systems. We encourage future work to further explore representational properties of LLMs and build efficiency-aware alignment strategies that require minimal external supervision.

\bibliography{natbib}
\bibliographystyle{unsrtnat}


\appendix

\section{Behavior Analysis}
\label{app:behavior}

We follow the taxonomy proposed by \cite{chen2025seal}, which categorizes reasoning steps into three distinct types:
\begin{itemize}
    \item Execution thoughts: These are the core reasoning steps in which the model directly analyzes the problem and performs the required computations or logical deductions in a step-by-step manner.
    \item Reflecting thoughts: These are metacognitive utterances where the model pauses to verify prior reasoning steps, check for possible errors, or express uncertainty about earlier conclusions.
    \item Transition thoughts: These are moments where the model explicitly shifts its reasoning direction, often adopting a new strategy mid-way through the solution.
\end{itemize}

This classification provides a lens for dissecting the functional composition of a reasoning trace and measuring how frequently models engage in potentially redundant or inefficient behaviors (e.g., unnecessary reflection or frequent transitions).

\end{document}